# Arithmetic Conditions give the Unpacking Strategy - A Numeral Decomposer for Incremental Unsupervised Grammar Induction

**Isidor Konrad Maier** *maier@b-tu.de*
**Matthias Wolff** *matthias.wolff@b-tu.de*
*Chair of Communication Engineering*
*Brandenburg University of Technology Cottbus - Senftenberg*
*DE-03046, Cottbus/Chóśebuz, Siemens-Halske-Ring 14, Deutschland*

**Abstract**

This paper presents a novel numeral decomposer based on arithmetic criteria. The criteria are not dependent on a base-10 assumption but only on *Hurford's Packing Strategy*. Hurford's Packing Strategy constitutes numerals by *packing* factors and summands to multiplicators. We found out that a numeral of value $n$ has a multiplicator larger than $\sqrt{n}$, a summand smaller than $\frac{n}{2}$ and a factor smaller than $\sqrt{n}$. Using these findings, the numeral decomposer attempts to detect and *unpack* factors and summand in order to reverse Hurford's Packing strategy. We tested its applicability for incremental unsupervised grammar induction in 273 languages. This way, grammars were obtained with sensible mathematical attributes that explain the structure of produced numerals. The numeral-decomposer-induced grammars are often close to expert-made and more compact than numeral grammars induced by a modern state-of-the-art grammar induction tool. Furthermore, this paper contains a report about the few cases of incorrect induced mathematical attributes, which are often linked to linguistic peculiarities like *context sensitivity*.

**Keywords:** Numeral Words, Hurdford's Packing Strategy, Numeral Decomposition, Incremental Grammar Induction, Context Sensitivity in Numerals

# 1 Introduction

## 1.1 Overview

For the purpose of grammar induction, a numeral decomposer is presented that is supposed to reverse Hurford's Packing Strategy (Hurford, 2011). The Packing Strategy - which is more in-depth explained in Section 2 - constitutes a numeral word by packing 0, 1 or 2 numerals to a base morpheme $M$, unless the numeral is just a digit word. In English, examples for base morphemes $M$ are 'teen', 'ty', 'hundred' and 'thousand'. The numerals packed to $M$ must be interpreted either by addition - in which case we call them summands - or by multiplication - in which case we call them factors. Example: For the English numeral 'two hundred sixty', two numerals are packed to the base morpheme 'hundred', 'two' as a factor and 'sixty' as a summand. Therefore, our numeral decomposition algorithm is supposed to unpack the subnumerals 'two' and 'sixty' when parsing the input $(260,$'two hundred sixty'), and therefore, the desired output would be '_ hundred _'$(2, 60)$.
In Section 3 we specify the objectives of the numeral decomposer.
The algorithm requires knowledge about the parsed numeral's number value, as well as a lexicon of other number-numeral pairs that allows to recognize subnumerals. The algorithm evaluates found subnumerals based on arithmetic criteria presented in Section 4. Based on the criteria, it decides whether or not to unpack them. If one assumed that the numeral must follow a base-10 system, criteria for decomposing are well known. One can calculate the decimal digits of the number value $n$ as $\lfloor \frac{n}{10^{k-1}} \rfloor \pmod{10}$ for $k = 1, 2, ...$ and detect the numeral words of the digits inside $n$'s numeral word, compare (Graben et al., 2019).

However, we do not assume a certain base system. Instead, we mainly rely on a finding that says that factors and summands of a numeral $N$ cannot have more than half of $N$'s value. Therefore, when $N$'s value is $n$, being $\leq \frac{n}{2}$ is a necessary criterion for a subnumeral of $N$ to be unpacked. Only this necessary criterion is used for the prototype numeral decomposer that we present in Section 5.1. It works in standard cases and it can fail if the numeral uses an unusual order of subnumerals, or if a certain critical subnumeral is not contained letter-by-letter, such as 'veinte' is not contained in 'veintiuno'. These details are described in Section 5.2. Extra unpacking criteria are established for an advanced numeral decomposer algorithm that fixes most errors that the prototype does. The advanced algorithm is presented in Section 5.3.
In Section 6 we present and discuss grammars induced by the numeral decomposer versions.

## 1.2 Related Work

Sproat (2022) wrote that neural networks nowadays still struggle to consistently form big numerals correctly. Therefore, numeral grammars are often programmed manually, see Akinadé and Odejobi (2014), Khamdamov et al. (2020), and Rhoda (2017). Numeral grammar induction is the way to automate the programming. The following numeral grammar induction approaches exist.

- Hammarström (2008) proposed a method that subdivides a set of numerals into $k$-sized clusters based on a similarity measure. Then generalizations can be made inside the clusters.
- Flach et al. (2000) made a proposal on how to automatically learn numeral grammars for finite-state transducers.
- Beim Graben et al. (2019) proposed that numerals may be added to a lexicon until a boundary is reached. Then a penalty signal arises that urges the learner increasingly to summarize several numerals in a generalization.

In a broader sense, the topic of numeral grammar induction is related to the linguistic theory of numerals and their computational modeling on the one hand and to general grammar induction of natural language on the other hand.
Regarding the linguistic theory of numerals, Brainerd (1966) collected seven studies on numeral grammars in different languages and attempted to draw general conclusions. Our work is mainly based on Hurford (2007). His theory is outlined in more detail in his book (Hurford, 2011). Ionin and Matushansky (2006) and Bhatt et al. (2005) describe similar theories on the structure of numerals. Derzhanski and Veneva (2018) give a summary of exceptional phenomena in the structures of numerals. Andersen (2004) discusses implications of the structure of numerals in various languages on the question whether or not all humans use an universal grammar. Mendia (2018) and Anderson (2019) discussed epistemic numbers, i.e. generalized number phrases like 'twenty-some'.
On the other hand, grammar induction is a wide field of research that refers to the process of learning formal grammars from data. It arose in the 1990s (Carroll & Charniak, 1992; Klein & Manning, 2001; Stolcke & Omohundro, 1994). A systematic and detailed review of the literature on unsupervised grammar induction till 2019 was performed by Muralidaran et al., 2021. Notably, only 1 out of 33 reviewed studies presented an incremental grammar learning method, namely Seginer (2007). Since 2018, significant advancements have been made, particularly with deep learning and neural models, leading to more effective and scalable grammar induction techniques. In particular, neural parametrization has proven to be useful in inducing PCFGs (Kim et al., 2019). Other neural approaches are Htut et al. (2018), Shen et al. (2019) and Drozdov et al. (2019). Two of the newest tools for natural language grammar induction are ShortcutGrammar (Friedman et al., 2022) and GITTA (Winters & Raedt, 2020). In this work, we use GITTA for another baseline comparison among Hammarström (2008) and Derzhanski and Veneva (2020). GITTA induces context-free grammars by using the Wagner-Fischer algorithm (Wagner & Fischer, 1974) to create common templates for similar expressions. Lately, heterogenous data that include vision or speech in addition to text have been discussed whether they can improve grammar induction, see Li et al. (2024) and Zhao et al. (2025).

## 1.3 Preliminary notations and wordings

In this subsection we establish our notation for numbers and numeral words.
Specific numbers are normally written with Hindu-Arab digits. For number variables, we use lower case letters. If $X$ is a numeral word, then $n(X)$ denotes the number of $X$. Often, we will also denote $n(X)$ by $X$'s lower-case letter $x$.
Specific numerals are written in quotation marks. For numeral or string variables, we use capital letters. If $x$ is any kind of number expression, then $N(x)$ denotes the numeral of $x$ in the language dealt with.[1] Often, we will also denote $N(x)$ by $x$'s upper-case letter $X$. By $\mathcal{N}$ we denote the set of numeral words of natural numbers $\mathbb{N}$ in the language dealt with.[2]

| Examples: | Numbers | Numerals |
|---|---|---|
| specific | $6$ | 'six' |
| variable | $x = 100$ | $X =$ 'hundred' |
| dependent | $n($'six'$\cdot X) = 600$ | $N(6 + x) =$ 'hundred and six' |

By '$\cdot$' we denote the concatenation of strings. We carry wordings for number relations over to numerals. The wordings

"Numeral $X$
is larger than / is smaller than / equals / is a divisor of / is a multiple of
numeral $Y$"

mean that the numbers $x$ and $y$ have the respective relation. This allows us to describe arithmetic between two numerals or between a numeral and a number with less effort.
Note that 'Numeral $X$ is larger than numeral $Y$' should not be interpreted as if $Y$ is a substring of $X$. In order to describe string relations between numerals we will only use the wordings 'is a substring/subnumeral/superstring/supernumeral of' or 'contains'/'is contained in'.
As mentioned before, the decomposer unpacks certain subnumerals out of a numeral. Suppose that in a numeral $X$ the subnumerals $U_1, ..., U_k$ are unpacked. This implies that $X = S_1 \cdot U_1 \cdot S_2 \cdot ... \cdot U_k \cdot S_{k+1}$ with strings $S_i$. Then we present the decomposition as

$$X = S_1_S_2_..._S_{k+1}(U_1, ..., U_k)$$

where the _ denote placeholders.
The term $S_1_S_2_..._S_{k+1}$ could be seen as an epistemic number expression that would be spoken $S_1\cdot$some$\cdot S_2\cdot$some...$\cdot S_{k+1}$, compare Mendia, 2018 and Anderson, 2019. In the following, $S_1_..._S_{k+1}$ is interpreted as a function of numeral words, defined on a domain $D \subset \mathcal{N}^k$:

$$S_1_..._S_{k+1} : D \to \mathcal{N}, (U_1, ..., U_k) \mapsto S_1 \cdot U_1 \cdot S_2 \cdot U_2 \cdot S_3 \cdot ... \cdot U_k \cdot S_{k+1} \quad (1)$$

We call $S_1_..._S_{k+1}$ the template or template function of the decomposition. Alternatively, when $x$ and $u_1, ..., u_k$ are the numbers of the numerals $X$ and $U_1, ..., U_k$, we can present the decomposition with the numbers as

$$x = S_1_S_2_..._S_{k+1}(u_1, ..., u_k).$$

We will usually use this notation because it is more compact.
The notation implies that $S_1_..._S_{k+1}$ can also be interpereted as a number function defined on $\mathbb{D} \subset \mathbb{N}^k$:

$$S_1_..._S_{k+1} : \mathbb{D} \to \mathbb{N}, (u_1, ..., u_k) \mapsto n(S_1 \cdot N(u_1) \cdot S_2 \cdot ... \cdot N(u_k) \cdot S_{k+1}) \quad (2)$$

Example:
In $X =$'twenty-seven thousand and two hundred and six' the subnumerals $N(27)$ and $N(206)$ are unpacked, so we present the decomposition as

$X =$ _ thousand and _('twenty-seven','two hundred and six') or
$27206 =$ _ thousand and _$(27, 206)$.

[1] If there is several ways how to spell $N(X)$ then we consider the different alternatives as seperate languages
[2] Numeral words only exist for a finite set of natural numbers in most languages, so $\mathcal{N}$ and $\mathbb{N}$ do not have a one-to-one correspondence.

The resulting numeral function is

$$_\text{ thousand and }_ : \{N(d) \mid d = 1, ..., 999\}^2 \to \mathcal{N}, (U_1, U_2) \mapsto U_1 \cdot \text{ thousand and } \cdot U_2$$

and the number function is

$$_\text{ thousand and }_ : \{1, ..., 999\}^2 \to \mathbb{N}, (u_1, u_2) \mapsto n(N(u_1) \cdot \text{ thousand and } \cdot N(u_2))$$

Humans understand easily that $n(N(u_1) \cdot$ thousand and $\cdot N(u_2))$ just means $1000 * u_1 + u_2$. For the general case, however, such an arithmetical equation is not trivial to find.

## 2 Axioms based on Hurford's Packing Strategy

First, we briefly summarize the explanation of the Packing Strategy from Hurford (2007):
Hurford says: "The Packing Strategy is a universal constraint on numeral systems. It applies very widely to developed numeral systems. It is not a truism, but exceptions are rare. The Packing Strategy operates in conjunction with a small set of phrase structure rules, which are shared by all developed numeral systems." These rules are given in figure 1.

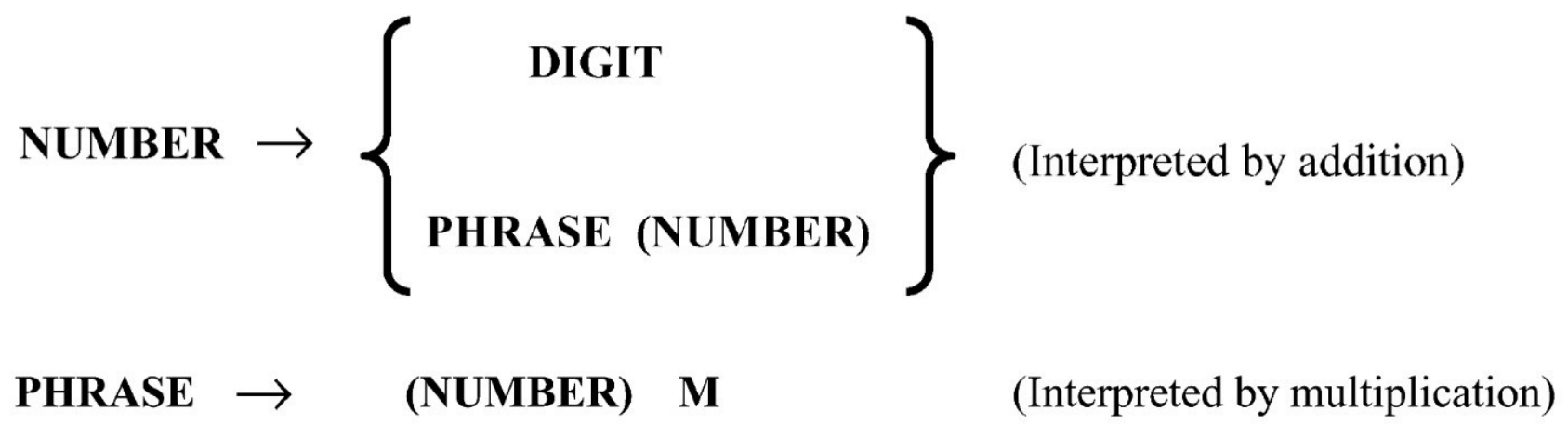

Figure 1: **Graphic originally from Hurford, 2007**
Explanation: Curly brackets indicate 'either/or' options, Parentheses indicate 'take-it-or-leave-it' optional choices. DIGIT is the category of basic lexical numerals, such as in English 'one',...,'nine'. M is the category of multiplicative base morphemes, such as in English 'ty-', 'teen', 'hundred', 'thousand' or 'million'.

Hurford also mentioned that

- in each rule, "the sister constituent of NUMBER must have the highest possible value".
- "the Packing Strategy says nothing about linear order, but only about the hierarchical dominance relationships between constituents of numeral expression".

We establish a new interpretation of the Packing Strategy based on the following axioms. An example of the English numeral $N(260)$ in Figure 2 shows the generation of the numeral according to Benford, as well as according to our reinterpretation.

**Axiom 1.** *The wording of each compound numeral $X$ implies a calculation of its number value $x$ as $x = fa * mu + su$, where $mu$ is the value of a multiplicative base morpheme from $M$ (see figure 1) and $fa$ and $su$ are the "factor" and "summand" numbers.*

**Justification.** *According to Hurford's Packing Strategy, $X$ is composed of* ***PHRASE (NUMBER)****, which should be interpreted as $phrase(+number)$. Further,* ***PHRASE*** *is composed* ***(NUMBER) M****, which means $(number*)m$.*
*Combined, we have $X =$* ***(NUMBER) M (NUMBER)****, which means $(number*)m(+number)$. After renaming the words, we have $X = (FA)MU(SU)$, which means $(fa*)mu(+su)$.*
*If both $FA$ and $SU$ exist in $X$, then our assumption is established.*
*If $FA$ is left out, then $X$ means $mu + su$. In this case we interpret it so that $X$ contains $FA$ as an empty string that implies the neutral factor $1$. This invisible string does not have an impact on the meaning, since $mu + su = 1 * mu + su$, so it does not hide information.*
*Likewise, if $SU$ is left out, then $X$ means $fa * mu$. In this case we interpret it so that $X$ contains $SU$ as an empty string that implies the neutral summand $0$. This empty string does not have an impact on the meaning, since $fa * mu = fa * mu + 0$.* □

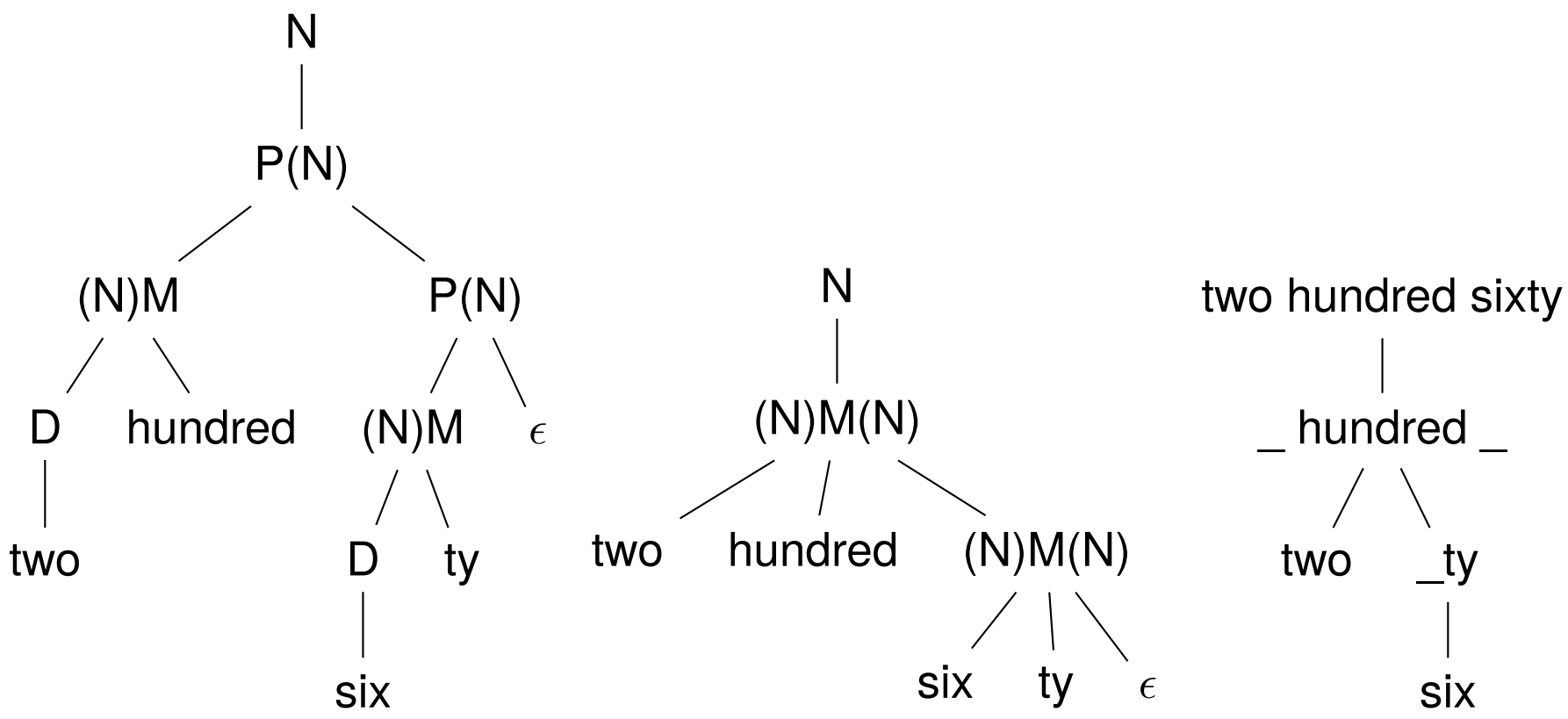

Figure 2: Packing the English numeral 'two hundred sixty'. *Left:* Generation following Hurford's phrase structure rules, compare Figure 1 with N=NUMBER, P=PHRASE and D=DIGIT. The Ns and Ms are bottom-up packed up. *Middle:* Following a version of Hurford's Packing Strategy that unites the rules $P \to P(N)$ and $P \to (N)M$ to $N \to (N)M(N)$. *Right:* The middle tree rephrased to a dependency tree using numeral functions that use the $M$ as the template and the $N$s as inputs. The tree suggests that the numeral decomposer should unpack $N(2)$ and $N(60)$ when parsing $N(260)$ and then when parsing $N(60)$ it should unpack $N(6)$.

As showcased in the justification of axiom 1, the supposed implication of a calculation $fa * mu + su$ does not mean that the corresponding numeral words $FA$, $MU$ and $SU$ are contained in $X$ letter-by-letter. If an implied subnumeral $FA$, $MU$ or $SU$ is not contained in $X$ letter-by-letter, we call it a **masked subnumeral**. Subnumerals can be masked for various reasons:
1) 'one hundred': As already mentioned, if $fa = 1$ or $su = 0$, then the implied subnumerals $FA$ and $SU$ can be left out because of redundancy, e.g., in English one simply says 'one hundred' instead of 'one hundred and zero'. In many other languages, the 'one' is also left out.
2) 'thirteen': Implied subnumerals can be subject to grammatical flexion, fusion with adjacent morphemes, or any other phenomenon that causes them to deviate from their standard form. E.g., in the English numeral 'thirteen', both the implied subnumerals 'three' and 'ten' are not letter-by-letter subnumerals because of that.
As a complement to Axiom 1, we assume:

**Axiom 2.** *Each subnumeral of $X$ that does not contain $MU$ is $FA$ or $SU$ or a subnumeral of those.*

Furthermore, we establish a basic assumption on the arithmetical relations between $fa$, $mu$ and $su$.

**Axiom 3.** *In the implied calculation $fa * mu + su$ of a numeral $X$, usually $fa < mu$ and $su < mu$.*

**Justification.** *If $fa \geq mu$, then $x \geq u^2$, so we suspect that a bigger multiplicative base morpheme number $mu$ could have been used, but "the sister constituent of NUMBER [$FA$] must have the highest possible value". If $su \geq mu$, we question why the numeral $X$ would not be formed as $FA'MUSU'$ with $n(FA') = fa + 1$ and $n(SU') = su - mu$.*

The established axioms are used in section 4 to prove unpack criteria that can distinguish the subnumerals $FA$ and $SU$—which are supposed to be unpacked—from other subnumerals.

# 3 Objectives

In this section we describe the practical objectives on the numeral decomposer algorithm. In Section 6 we describe to what degree these objectives have been achieved.
We had initially stated that the goal is to design a numeral decomposer that mimics Hurford's Packing Strategy, i.e., the factor word $FA$ and the summand word $SU$ should be unpacked. Due to the practicality of the application in 254 natural languages, we state our objectives in a more pragmatic way. Given a flat list of number-numeral pairs

$$\{(1,\text{'one'}),(2,\text{'two'}),...(4,\text{'four'}),...(14,\text{'fourteen'})...\},$$

each numeral is distributed to a template by the numeral decomposer. While for basic numerals like 'one', 'two', or 'four' the template will be the word itself, a complex numeral like 'fourteen' can be distributed to '_teen'. In either case, the original numeral can be reconstructed from the template. In the cases of 'one, 'two' and 'four' the reconstruction is trivial, while in the case of '_teen' the template has to be combined with another template 'four' in order to reconstruct 'fourteen'. Since the templates can collectively reconstruct the original numerals, the set of templates can constitute a lexicon that—with given grammar rules—can generate the original list. So, the original list can be replaced by the lexicon of templates.
The lexicon of templates is smaller than the original list, because several numerals can share one template. E.g., if the numbers $14, 16, 17, 19$ are decomposed as

$$14 = \text{_teen}(4), 16 = \text{_teen}(6), 17 = \text{_teen}(7) \text{ and } 19 = \text{_teen}(9),$$

then four entries of the original list are replaced by one entry in the lexicon of templates. As described in section 1.3, the template '_teen' can then be described as a function that operates in the domain $\{4, 6, 7, 9\}$. And a template '_ hundred and _' may comprise $9 * 99$ numerals in one entry. Overall, a reduction of lexicon is achieved if numerals are decomposed in a uniform way.

**Objective 1** (Compactness)**.** *The numeral decomposer should produce as few different templates as possible.*

However, templates should not be too uniform. Numerals that share a common template should have a reasonable relation. The English numerals 'twenty-one' and 'twenty-seven thousand' could both be distributed to one template, which may be 'twenty-_' or '_-_', even if their mathematical relation seems unreasonable. We define a reasonable relation in the following way:

**Objective 2** (Correctness)**.** *The functional equation of each template function must be affine linear.*

## 4 Unpack Criteria

Based on our new interpretation of Hurford's Packing Strategy, we establish unpack criteria, i.e., criteria that distinguish the factor $FA$, the summand $SU$ and any sub-subnumerals of the two on the one hand from $MU$ and subnumerals containing $MU$ on the other hand. The unpack criteria are mathematically proven under the following working assumption.
**Working assumption:** *If a numeral $N$ contains another numeral $N'$ then $n' \leq n$.*
We comment on its validity at the end of this chapter.

**Unpack criterion 1** (Necessary unpack criterion)**.** *Let $X$ be a numeral. Let $S$ be a subnumeral of $X$ that is contained in $X$'s factor word $FA$ or in $X$'s summand word $SU$. Then $2 * s < x$*

*Proof.* We assume that $S$ being contained in $FA$ or $SU$ implies $s \leq fa$ or $s \leq su$.
If $s \leq fa$, then

$$2 * s \leq 2 * fa \leq mu * fa \leq fa * mu + su = x.$$

Thus, we have $2 * s < x$ unless the numeral system is base 2 and $mu = 2$ and $su = 0$.
If $s \leq su$, then

$$2 * s \leq 2 * su = su + su < mu + su \leq fa * mu + su = x.$$

Thus, in either case $2 * s < x$. □

This criterion alone was sufficient to construct a prototype numeral decomposer, see I. Maier and Wolff, 2022.

**Unpack criterion 2** (Sufficient unpack criterion)**.** *Let $X$ be a numeral. Let $S$ be a subnumeral of $X$ that satisfies $s^2 \leq n$. Then $S$ is either contained in $X$'s factor word $FA$ or in $X$'s summand word $SU$.*

*Proof.* If $S$ is neither contained in $FA$ nor $SU$, then by axiom 2, $S$ contains $MU$. Therefore we assume that $s \geq mu$. Then, since $fa < mu$ and $su < mu$, we have

$$s^2 \geq mu^2 = mu * mu = (mu-1) * mu + mu > (mu-1) * mu + (mu-1) \geq fa * mu + su = x.$$

Hence, $S$ must be contained in either $FA$ or $SU$. □

The criteria can be used to decide whether or not to unpack subnumerals. The criteria do not refer to $FA$ and $SU$ directly but also to all sub-subnumerals of those. Thus, in order to use the criteria, one should look for the maximal subnumerals that still match the criteria.
The present criteria leave a gap for subnumerals valued between $\sqrt{x}$ and $\frac{x}{2}$, for which further unpack criteria are needed.
Next, we add a classification for subnumerals that are supposed to be unpacked despite not satisfying unpack criterion 2:

**Unpack criterion 3.** *Let $X$ be a numeral. Let $S$ be a subnumeral of $X$ that does not contain $X$'s multiplicator word $MU$, but let $s^2 > x$. Then $S$ is $SU$ or contained in $SU$.*

*Proof.* By Axiom 2, $S$ must be equal or contained in either $FA$ or $SU$. If it is contained in $FA$, we assume $s \leq fa$, so

$$s^2 \leq fa^2 < fa * mu \leq fa * mu + su = x.$$

Thus $S$ must be contained in $SU$. □

The present criteria can be summarized as:

| | |
|---|---|
| $s < \sqrt{n}$ | $\Rightarrow S$ is (contained in) $FA$ or $SU$ |
| $\sqrt{n} < s < n/2$ | $\Rightarrow S$ can be (contained in) $SU$ |
| $n/2 < s$ | $\Rightarrow S$ is not (contained in) $FA$ or $SU$ |

Subnumerals valued between $\sqrt{n}$ and $n/2$ remain undecidable up to this point. Examples for such yet undecidable subnumerals are abundant. E.g. in English,$N(26) =$ 'twenty-six' has the summand word $SU = N(6)$, but $6^2 > 26$. So, it is not yet decidable, whether $N(6)$ is a summand or not. The algorithm cannot exclude that the numeral uses base 6 with $x = 4 * 6 + 2$, so $N(6)$ would be the multiplicator. The given counterexample can be generalized to any base 10 system. For a working decomposer we should close the gap of decision. We could not find a definitive solution. Instead, we use a leaky criterion based on the idea that $su$ is usually not a divisor of $x$.

**Unpack criterion 4** (Leaky criterion)**.** *Let $X$ be a numeral to be interpreted as $fa * mu + su$. Let $S$ be a subnumeral of $X$. Assume that*

- *$N(fa * mu)$ is masked in $X$,*
- *$x/2 > s > \sqrt{x}$ and*
- *$S$ has no subnumerals.*

*Then we assume that $S$ is a subnumeral of $SU$ if and only if $s \nmid fa * mu$.*

**Justification.** *$\Leftarrow$: Given that $s^2 > x$, by criterion 3 $S$ cannot be contained in $FA$. If $s \nmid fa * mu$, then $S$ cannot be $MU$ either. $S$ can be 1)$N(fa' * mu)$ with a subnumeral $FA'$ of $FA$ with $fa' \nmid fa$ or 2)$N(mu + su')$ with a subnumeral $SU'$ of $SU$ or 3)$SU$. Any $N(fa' * mu)$ or $N(mu + su')$ is unlikely to be a subnumeral of $X$ given that $N(fa * mu)$ is masked. Also, they likely have subnumerals unlike $S$. Hence, we assume that $S$ is a subnumeral of $SU$.*
*$\Rightarrow$: In this direction, we claim that there is no $x = fa * mu + su$ with $N(fa * mu)$ being masked, $SU$ having no subnumerals, $su^2 > x$ and $su \mid fa * mu$. Although this claim is not generally true, we explain why exceptions are rare and showcase which amount of coincidences they require. First, it is rare to have $N(fa * mu)$ masked, especially at higher numbers with 3 digits[3] or more. And even if a 3-digit numeral would have $N(fa * mu)$ masked, then it would still require $N(su)$ to have no subnumerals, which often means that $su$ is small, so it is unlikely that $su^2 > x$.*

[3]'Digit' does not necessarily refer to base 10 digits here, but more generally to the coefficients $c_i$ in base $b$ representation $c_0 * b^0 + c_1 * b^1 + ...$

*For 2-digit numbers, $N(su)$ is 1-digit, so there are few possibilities to construct an exception. In base 10, the only pairs $(fa, su)$ that satisfy the arithmetic properties $su \mid fa * mu$ and $su^2 > fa * 10 + su$ are $(1, 5)$ and $(4, 8)$. So, if in English $N(48)$ would be 'fortaj-eight', while $N(40)$ would still be 'fort**y**', then 'eight' can be suspected as a multiplicator, because $8 \mid 48$. Vigesimal base 20 systems offer more space for exceptions. Arithmetically, with $mu = 20$ they are possible if $(fa, su) \in$*

$$\{(1, 10), (2, 10), (3, 10), (4, 10), (3, 12), (6, 12), (7, 14), (3, 15), (6, 15), (9, 15), (4, 16), (8, 16), (9, 18)\}.$$

*So, whenever a vigesimal numeral has its subnumeral $N(fa * mu)$ masked, an exception could occur. It occurs in French where the numerals $N(4 * 20 + k)$ for $k = 1, ..., 19$ are spelled 'quatre-vingt-$\cdot N(k)$' and do not contain $N(4 * 20) =$ 'quatre-vingt**s**' letter-by-letter. The numerals $N(4 * 20 + 10) =$ 'quatre-vingt-dix' and $N(4 * 20 + 16) =$ 'quatre-vingt-seize' also fulfill the arithmetic requirements, and $N(10) =$ 'dix' and $N(16) =$ 'seize' also do not have subnumerals. Posing an exception to unpack criterion 4 leeds to $N(10)$ and $N(16)$ being interpreted as multiplicators, as will be seen later.*

The use case for unpack criterion 4 may not seem obvious, but in Section 5.3 one can find a situation for which it is tailor-made. The leakiness of unpack criterion 4 is not as problematic, as its failure can only cause a small lack in generalization (Objective 1) rather than a problematic overgeneralization (Objective 2).
**Comment on working assumption:** Generally, it is possible that a numeral $N$ is smaller than its subnumeral. Derzhanski and Veneva (2018) mention two possibilities, namely when $N$'s structure involves overcounting SOURCE or subtraction. These possibilities lead to logical violations of the unpacking criteria. However, the unpack criteria keep their present validity for the main cases $S \in \{FA, MU, SU\}$ and since subtraction and overcounting only allow $n'$ to be lightly larger than $n$, there are few possibilities for logical violation. Such violation did not practically harm our grammar induction tests.

# 5 Numeral decomposer algorithm

## 5.1 Prototype

First, we had discovered unpack criterion 1 and noticed that it alone can drive a decent decomposition algorithm. Since it is just a necessary but not sufficient criterion, it may unpack subnumerals larger than $FA$ and $SU$. Specifically, $mu$ and $mu + su$ are often $< n/2$. In the prototype algorithm 1 we circumvent this issue by setting a checkpoint so that $N(mu)$ and $N(mu + su)$ are never tested on the criterion in the first place.
For illustration, we describe the decomposition of the complex English numeral $N(27001) =$ 'twenty-seven thousand and one'.
The $end$- and $start$-loops (lines 3,4) make the code check substrings of 'twenty-seven thousand and one' in the order 't', 'tw', 'w', 'twe', 'we', 'e', 'twen', 'wen'... At $[start{:}end] = [0{:}6]$, the subnumeral $N(20) =$'twenty' is found (ln. 6). Since $20 < 27001/2$ (ln. 7), it gets unpacked (ln. 8). Then the $start$ loop gets broken (ln. 13), so the next substring to check is 'twenty-' at $[start{:}end] = [0{:}7]$ rather than 'wenty' at [1:6]. At [0:12], $N(27) =$'twenty-seven' is found (ln. 6). Since $27 < 27001/2$ (ln. 7), it is also unpacked (ln. 8) and its previously unpacked sub-subnumeral $N(20) =$'twenty' is un-unpacked (ln. 9). Moreover, $start$-loop is broken (ln. 13), so the next substring to check is at $[0{:}13]$ 'twenty-seven ' and the algorithm will never see the $N(7) =$'seven' at $[7{:}12]$. With lines 9 and 13 we make sure that the factor word $N(27)$ is unpacked in one rather than having $N(20)$ and $N(7)$ being unpacked separately. At [0:21], $N(27000) =$'twenty-seven thousand' is discovered. Since $27000 \not< 27001/2$ (ln. 7,10), it is not unpacked but the $start$ loop is broken (l. 13), so the algorithm will continue at $end = 22$. This way, it is avoided that the algorithm finds $N(7000)$ at [7:21] or $N(mu) = N(1000)$[4] at $[13{:}21]$. If it would find $N(7000)$ or $N(1000)$, it would unpack it, since it is $< 27001/2$. Also, the checkpoint $cp$ is reset to $21$ (ln. 11), which makes the $start$ loop no longer check substrings with $start < 21$ (ln. 4). This way, we avoid that $N(7001)$ at $[7{:}29]$ or $N(1001)$ at $[13{:}29]$ may be found and unpacked. Instead, it will just find

[4]Officially, $N(1000)$ is 'one thousand' in English. However, we show that it even works if it was just spelled 'thousand'. Same holds for 'thousand and one'.

**Algorithm 1** Prototype algorithm using Python syntax. $cp$ stands for checkpoint. Function isNumeral returns $True$ iff the input string is a grammatically formed numeral word based on an available lexicon. $n(substring)$ is the number value of the numeral $substring$. Instruction 'Unpack $substring$' adds $substring$ to a list of unpacked subnumerals. An 'Un-unpack...' instruction removes an entry from the list of unpacked subnumerals.

```
1: Decompose numeral
2: cp ← 0
3: for end in range(length(numeral)) do:
4:     for start in range(cp:end) do:
5:         substring ← numeral[start:end]
6:         if substring isNumeral then:
7:             if 2 * n(substring) < n(numeral) then:
8:                 Unpack substring
9:                 Un-unpack sub-substrings of substring that were unpacked before
10:            else:
11:                cp ← end
12:            end if
13:            break start-loop
14:        end if
15:    end for
16: end for
```

$N(1)$ at [26:29] (ln. 6) and unpack it (ln. 7). The algorithm terminates then (ln. 13) and - as intended - has unpacked the subnumerals $FA = N(27)$ and $SU = N(1)$.
We systematically describe numeral decompositions in the following form.

```
Language: English   Number: 27001 = 27*1000+1
Index:        0 1 2 3 4 5 6 7 8 9 10  12    15        20        25    28
Numeral:      t w e n t y - s e v e n   t h o u s a n d   a n d   o n e
Subnumerals:|---N(20)---| |-----------------N(7001)-----------------|
            |--------N(27)----------| |-----------N(1001)-----------|
                          |--N(7)---| |----N(1000)----|         |-N(1)|
            |-----------------N(27000)---------------|
                          |---------N(7000)----------|
Desired decomposition: _ thousand and _(27,1)
```

'twenty-seven thousand and one' decomposed by prototype algorithm 1

| $[start{:}end]$: | [0:6] | [0:12] | [0:21] | [26:29] |
|---|---|---|---|---|
| Subnumeral: | $N(20)$ | $N(27)$ | $N(27000)$ | $N(1)$ |
| Criterion: | $< 27001/2$ | $< 27001/2$ | $\not< 27001/2$ | $< 27001/2$ |
| Checkpoint: | 0 | 0 | $0 \to 21$ | 21 |
| Unpacked: | $\{20\}$ | $\{27\}$ | $\{27\}$ | $\{27, 1\}$ |
| References: | ln.7,8 | ln.7-9 | ln.10,11 | ln.7,8 |

⇒_thousand and _$(27, 1)$

A numeral description form is provided that includes the numeral's language, number value, and desired decomposition and visualizes **all** its subnumerals. A 0-based letter index scale facilitates referencing between $[start{:}end]$ values and substrings.To describe the decomposition process, we use tables that describe the algorithm's behaviour at every time when a subnumeral is found. The present example of the decomposition of 'twenty-seven thousand and one' showcases that the prototype algorithm unpacks exactly $FA$ and $SU$ when all of the following 4 conditions are true.

1. The subwords $FA$, $MU$ and $SU$ are arranged in the order $FA \cdot MU \cdot SU$
2. $N(fa * mu)$ is not masked. (Otherwise, the checkpoint is not reset properly, so $N(mu + su)$ may be unpacked instead of $N(su)$)
3. $N(mu)$ ends at the same letter as $N(fa * mu)$ or is masked. (Otherwise, $N(mu)$ may be unpacked)

4. $N(fa)$ and $N(su)$ are not masked. (Otherwise, they cannot be found and unpacked.)

Under these conditions, the algorithm perfectly reverses Hurford's Packing Strategy.
The following example of the Tsez numeral $N(86) =$'uynoquno iłno' shows that the prototype works fine with numerals that follow a non-base-10 system:

```
Language: Tsez   Number: 86 = 4*20+6
Index:       0       4      8 9      13
Numeral:     u y n o q u n o   i ł n o
Subnumerals:|-N(4)--|-N(20)-| |-N(6)--|
            |----N(80)------|
                    |-----N(26)-------|
Desired decomposition: _quno _(4,6)
```

'uynoquno iłno' decomposed by prototype algorithm 1

| $[start{:}end]$: | [0:4] | [0:8] | [9:13] | |
|---|---|---|---|---|
| Subnumeral: | $N(4)$ | $N(80)$ | $N(6)$ | |
| Criterion: | $< 86/2$ | $\geq 86/2$ | $< 86/2$ | |
| Checkpoint: | 0 | $0 \to 8$ | 8 | ⇒_quno _$(4,6)$ |
| Unpacked: | $\{4\}$ | $\{4\}$ | $\{4,6\}$ | |
| References: | ln.7~8 | ln.10,11 | ln.7~8 | |

The resulting template '_quno ' is obtained analogously when parsing any other Tsez numeral $N(a*20+b)$ for $(a,b) \in \{2,3,4\} \times \{1,...,19\}$, so it constitutes a proper functional equation $(x_1,x_2) \mapsto 20x_1 + x_2$.
In I. Maier and Wolff, 2022 we already published the present prototype decomposition algorithm and showed that it is surprizingly well-rounded. However, it still has systematic errors, which we show in the following analysis.

## 5.2 Errors of prototype algorithm

In the last subsection we showed that errors do not appear if the parsed numeral has a generic $FA \cdot MU \cdot SU$ order and no subnumerals are masked. In this section, we show what errors can appear otherwise and what issues they can cause with respect to the objectives stated in Section 3. This is not a complete analysis of what errors could theoretically appear, but only a summary of what errors we found in the database.

**Masked subnumerals:** Often, subnumerals get assimilated by the surrounding word, so they are not letter-by-letter subnumerals. For example, in 'thirty' the implied subnumerals 'three' and 'ten' are both masked, so they cannot be found letter-by-letter. We do not aim to find any summand or factor in a numeral if it is masked. This would be a task of pattern recognition. However, masked subnumerals can also cause a factor or summand not to be unpacked, even if the factor or summand itself is not masked, as the example of Spanish $N(25) =$'veinticinco' shows. Here, since $FA \cdot MU = N(20) =$'veinte' is masked, the checkpoint $cp$ is not moved early enough.

```
Language: Spanish   Number: 25 = 20+5
Index:       0           6       10
Numeral:     v e i n t i c i n c o
Subnumerals:           |---N(5)--|
Desired decomposition: veinti_(5)
```

'veinticinco' decomposed by prototype algorithm 1

| $[start{:}end]$: | [0:11] | |
|---|---|---|
| Subnumeral: | $N(25)$ | |
| Criterion: | $\not< 25/2$ | |
| Checkpoint: | $0 \to 9$ | ⇒veinticinco() |
| Unpacked: | $\{\}$ | |
| References: | ln.10,11,13 | |

The algorithm only enters the if-clause in line 6 for the first time at $[0{:}11]$ with the total numeral $N(25) =$'veinticinco'. Nothing is unpacked because $25 > 25/2$ (ln. 7) and the $start$-loop gets broken (ln. 13), causing the algorithm not to find $N(5) =$'cinco' at all.
The same issue concerns all Spanish numerals from $N(21)$ to $N(29)$.
We state: **An error can be caused by $N(fa * mu)$ being masked.**

**Order of subnumerals**: According to Hurford, the subnumerals $FA$, $MU$ and $SU$ can be in a different order, since the rules in Figure 1 only represent a hierarchy.
Order $SU \cdot FA \cdot MU$: This order of subnumerals does not generally cause undue decomposition, as the following example showcases:

```
Language: Upper-Sorbian     Number: 61 = 1+6*10
Index:        0         5 6       10       15
Numeral:      j ě d y n a š ě s ć d ź e s a t
Subnumerals:|---N(1)--| |--N(6)-|
                        |-------N(60)-------|
Desired decomposition: _a_dźesat(1,6)
```

'jedynašěsćdźesat' decomposed by prototype algorithm 1

| $[start{:}end]$: | [0:5] | [6:10] | [0:16] |
|---|---|---|---|
| Subnumeral: | $N(1)$ | $N(6)$ | $N(61)$ |
| Criterion: | $< 61/2$ | $< 61/2$ | $\not< 61/2$ |
| Checkpoint: | 0 | 0 | $0 \to 16$ |
| Unpacked: | $\{1\}$ | $\{1,6\}$ | $\{1,6\}$ |
| References: | ln. 7,8 | ln. 7,8 | ln. 10,13 |

⇒_a_dźesat$(1,6)$

Similar cases are dealt with in Somali, Lower-Sorbian, Slovene, and many Germanic languages. The order $FA \cdot SU \cdot MU$ would be decomposed in the same way, but we have not found any real examples for it.
Order $MU \cdot FA \cdot SU$: This order of subnumerals can cause $MU$ to be unpacked instead of $FA$ as the following example showcases:

```
Language: Nyungwe      Number: 34 = 10*3+4
Index:        0   2         7     10      14          20  22
Numeral:      m a k ´ u m i   m a t a t u   n a   z i n a i
Subnumerals:     |--N(10)--|     |--N(3)-|             |-N(4)|
            |-----------N(30)-----------|
Desired decomposition: mak´umi ma_ na zi_(3,4)
```

'mak´umi matatu na zinai' decomposed by prototype algorithm 1

| $[start{:}end]$: | [2:7] | [0:13] | [19:22] |
|---|---|---|---|
| Subnumeral: | $N(10)$ | $N(30)$ | $N(4)$ |
| Criterion: | $< 34/2$ | $\not< 34/2$ | $< 34/2$ |
| Checkpoint: | 0 | $0 \to 13$ | 13 |
| Unpacked: | $\{10\}$ | $\{10\}$ | $\{10,4\}$ |
| References: | ln. 7,8 | ln. 7,8,9,13 | ln. 7,8 |

⇒ma_ matatu na zi_$(10,4)$

The problem is that the subnumeral $n(fa * mu)$—that usually is the first to be $\geq n/2$—is not finished by $N(mu)$ but by $N(fa)$, which lets the prototype algorithm confuse $FA$ with $MU$.
If $FA$ is a compound numeral the algorithm may only interpret an initial part of $FA$ as $MU$ as the following example showcases.

```
Language: Makhuwa      Number: 60 = 10*(5+1)
Index:        0                9         14       18    21
Numeral:      m i l o k o   m i t h a n u   n a   m o s a
Subnumerals:|-----------N(50)-----------|       |--N(1)-|
                              |---N(5)--|
                              |-----------N(6)----------|
Desired decomposition: miloko mi_(6)
```

'miloko mithanu na mosa' decomposed by prototype algorithm 1

| $[start{:}end]$: | [0:14] | [18:22] |
|---|---|---|
| Subnumeral: | $N(50)$ | $N(1)$ |
| Criterion: | $\not< 60/2$ | $< 60/2$ |
| Checkpoint: | $0 \to 14$ | 14 |
| Unpacked: | $\{\}$ | $\{1\}$ |
| References: | ln. 10,11,13 | ln. 7,8 |

$\Rightarrow$miloko mithanu na _$(1)$

The same issue would arise whenever $MU$ stands before $FA$, also in orders $MU \cdot SU \cdot FA$ and $SU \cdot MU \cdot FA$. However, we did not find any natural numerals arranged like this.
We state: An error can be caused by $MU$ appearing before $FA$.
We can generalize the statement to: **An error can be caused by $N(mu)$ ending before $N(fa * mu)$**
An example is found in Suomi (Finnish, fi).

```
Language: Suomi     Number: 201 = 2*100+1
Index:       0        5      9 10    13
Numeral:     k a k s i s a t a a y k s i
Subnumerals:|---N(2)--|-N(100)| |--N(1)-|
            |-------N(200)------|
Desired decomposition: _sataa_(2,1)
```

'kaksisataayksi' decomposed by prototype algorithm 1

| $[start{:}end]$: | [0:5] | [5:9] | [0:10] | [10:14] |
|---|---|---|---|---|
| Subnumeral: | $N(2)$ | $N(100)$ | $N(200)$ | $N(201)$ |
| Criterion: | $< 201/2$ | $< 201/2$ | $\not< 34/2$ | $< 34/2$ |
| Checkpoint: | 0 | 0 | $0 \to 10$ | 10 |
| Unpacked: | $\{2\}$ | $\{2,100\}$ | $\{2,100\}$ | $\{2,100,1\}$ |
| References: | ln. 7,8 | ln. 7,8 | ln. 10,11 | ln. 7,8 |

$\Rightarrow$__a_$(2,100,1)$

In this case, both $N(fa)$ and $N(mu)$ get unpacked, as neither is right at the end of $N(fa * mu)$.
Overall, we have classified two causes of error:

- **Cause 1: Masked** $FA \cdot MU$ ($N(fa * mu)$ is no subnumeral).
- **Cause 2: Early** $MU$ ($N(mu)$ ends before $N(fa * mu)$ ends)

These causes can lead to two different sorts of problems:

- Sort 1: $FA$ or $SU$ not getting unpacked, despite not being masked.
- Sort 2: $MU$ getting unpacked.

Either cause can lead to either sort of problem. For each combination, an example numeral word is given in the following table. Each example is explained in this subsection.

| | $FA$ or $SU$ not unpacked | $MU$ unpacked |
|---|---|---|
| $FA \cdot MU$ masked | 'veinticinco' | 'quatre-vingt-seize' |
| Early $MU$ | 'mak´umi matatu na zinai' | 'kaksisataayksi' |

Problems of sort 1 lead to issues with lexicon size (Objective 1). Whenever a $FA$ or $SU$ is not unpacked in a numeral $X = N(fa * mu + su)$, the numeral $X$ cannot be identified with similar numerals like $N(fa' * mu + su)$ or $N(fa * mu + su')$, so $X$ would need its own template.
Problems of sort 2 can cause wrong functional equations (Objective 2) due to undue generalization. As already mentioned in Section 3, if in English $N(21) - N(29)$ got generalized with $N(27000)$ to a single function twenty-_ with input set $\{1, ..., 9, 7000\}$, a correct affine linear functional equation would not exist.

## 5.3 Advanced Algorithm

In this subsection we explain how algorithm 2 solves issues of the prototype algorithm. In the following subsections, we show how the added lines 12-32 enhance lexicon size reduction and how the added lines 38-60 avoid overgeneralization.

**Algorithm 2** Advance numeral decomposition algorithm. In the new lines 12-32, further criteria are added under which subnumerals can get unpacked. The new lines 38-60 are tests based on which unpacked multiplicators are detected and un-unpacked.

```
 1: Decompose numeral
 2: cp ← 0
 3: for end in range(length(numeral)) do:
 4:     for start1 in range(cp:end) do:
 5:         substring ← numeral[start1:end]
 6:         if substring isNumeral then:
 7:             if 2 * n(substring) < n(numeral) then:
 8:                 Unpack substring
 9:                 Un-unpack sub-substrings of substring that were unpacked before
10:             else:
11:                 cp ← end
12:                 for start2 in range(start1 + 1, end) do:
13:                     substring2 ← numeral[start2:end]
14:                     if substring2 isNumeral then:
15:                         if n(substring2)^2 ≤ n(numeral) then:
16:                             Unpack substring2
17:                             Un-unpack sub-substrings of substring2
18:                             cp ← start2
19:                             maybeUnpack ← None
20:                             break start2-loop
21:                         else if n(substring2) ∤ n(substring) and n(substring2) * 2 < n(numeral) then:
22:                             maybeUnpack ← substring2
23:                             maybeCP ← start2
24:                         else:
25:                             maybeUnpack ← None
26:                         end if
27:                     end if
28:                 end for
29:                 if maybeUnpack ≠ None then:
30:                     Unpack maybeUnpack
31:                     cp ← maybeCP
32:                 end if
33:             end if
34:             break start1-loop
35:         end if
36:     end for
37: end for
38: maybeMU ← value-largest unpacked subnumeral
39: otherUnpac ← {unpacked subnumerals} \ {propMU}
40: if length(otherUnpac) = 1 then:
41:     if n(maybeMU) + n(otherUnpac) = n(numeral) then:
42:         Un-unpack maybeMU
43:     else if n(maybeMU) * n(otherUnpac) = n(numeral) then:
44:         Un-unpack maybeMU
45:     end if
46: else if length(otherUnpac) = 2 then:
47:     if n(otherUnpac[0]) * n(maybeMU) + n(otherUnpac[1]) = n(numeral) then:
48:         Un-unpack maybeMU
49:     else if n(otherUnpac[1]) * n(maybeMU) + n(otherUnpac[0]) = n(numeral) then:
50:         Un-unpack maybeMU
51:     end if
52: else if length(otherUnpac) > 2 then:
53:     for unpacked in otherUnpac do:
54:         maybeFA ← unpacked
55:         maybeSUs ← otherUnpac \ {maybeFA}
56:         if maybeFA * maybeMU + ∑(maybeSUs) = n(numeral) then:
57:             Un-unpack maybeMU
58:         end if
59:     end for
60: end if
```

### 5.3.1 Dealing with lexicon reduction

In this section we explain the lines 1-37 in the advanced algorithm 2. These code lines are built out of algorithm 1 and the added lines 12-32. The added lines deal with errors from algorithm 1 where $FA$ or $SA$ did not get unpacked despite not being masked, such as in the cases of 'veinticinco' and 'mak´umi matatu na zinai'. Fixing these errors enhances desired generalizations of words, where $FA$ or $SU$ can be replaced by other factor or summand words. In this way, the lexicon size can be reduced. We present their use cases in the following examples.
Recall the decomposition process of 'mak´umi matatu na zinai' with algorithm 1:

`'mak´umi matatu na zinai' decomposed by prototype algorithm 1`

| $[start{:}end]$: | [2:6] | [0:13] | [19:22] |
|---|---|---|---|
| Subnumeral: | $N(10)$ | $N(30)$ | $N(4)$ |
| Criterion: | $< 34/2$ | $\not< 34/2$ | $< 34/2$ |
| Checkpoint: | 0 | $0 \rightarrow 13$ | 13 |
| Unpacked: | $\{10\}$ | $\{1\}$ | $\{10, 4\}$ |
| References: | ln. 7,8 | ln. 10,11,13 | ln. 7,8 |

$\Rightarrow$ma_ matatu na zi_$(10, 4)$

The factor 'tatu' did not get unpacked due to early $MU$ (Cause 2). In order to resolve the issue, in line 12, we open a second $start2$ loop that looks for more substrings $substring2$ that end at the current $end$ (ln. 13), so the factor 'tatu' can be found at all. In line 15 we check $substring2$ for unpack criterion 2. Since $3^2 \leq 34$, the if clause is entered and $substring2 = N(3) =$'tatu' is unpacked in line 16. In detail, lines 1-37 of algorithm 2 process 'mak´umi matatu na zinai' the following way:

`'mak´umi matatu na zinai' decomposed by ln. 1-37 of advanced alg. 2:`

| $[start{:}end]$: | [2:6] | [0:13] | [9:13] | [19:22] |
|---|---|---|---|---|
| Subnumeral: | $N(10)$ | $N(30)$ | $N(3)$ | $N(4)$ |
| Criterion: | $< 34/2$ | $\not< 34/2$ | $\leq \sqrt{34}$ | $< 34/2$ |
| Checkpoint: | 0 | $0 \rightarrow 13$ | $13 \rightarrow 9$ | 9 |
| Unpacked: | $\{10\}$ | $\{10\}$ | $\{10, 3\}$ | $\{10, 3, 4\}$ |
| References: | ln. 7,8 | ln. 10,11 | ln. 15,16 | ln. 7,8 |

$\Rightarrow$ma_ ma_ na zi_$(10, 3, 4)$

We will show in subsection 5.3.2, that $N(10) =$'k´umi' will still get un-unpacked by the algorithm lines 38-60. In this way the desired decomposition $34 =$mak´umi ma_ na zi_$(3, 4)$ will be obtained finally. Note the following details:

1. In line 15, we use the sufficient unpack criterion 2 rather than the necessary unpack criterion 1 to avoid unpacking multiplicator words. If we used criterion 1 instead, errors would appear frequently in basic cases, like in English 'two hundred and one'. When $N(100) =$'hundred'[5] is found, criterion 1 would unpack it, while criterion 2 does not, since $\sqrt{201} < 100 < 201/2$.

2. In line 18, the checkpoint is reset from the current $end$ to the current $start2$. This becomes important in examples such as the following where the order is $MU \cdot FA \cdot SU$ and $FA$ is a composed numeral.

```
Language: Makhuwa     Number: 60 = 10*(5+1)
Index:       0              9         14      18    21
Numeral:     m i l o k o   m i t h a n u   n a   m o s a
Subnumerals:|-----------N(50)-----------|       |--N(1)-|
                            |---N(5)--|
                            |-----------N(6)----------|
Desired decomposition: miloko mi_(6)
```

[5]Here we assume that $N(100) =$'hundred' rather than 'one hundred'. Otherwise, the example works similar with $N(201)$ in Deutsch or various other languages.

'miloko mithanu na mosa' decomposed by advanced algorithm 2:

| $[start{:}end]$: | [0:14] | [9:14] | [9:22] | |
|---|---|---|---|---|
| Subnumeral: | $N(50)$ | $N(5)$ | $N(6)$ | |
| Criterion: | $\not< 60/2$ | $\leq \sqrt{60}$ | $< 60/2$ | ⇒miloko mi_$(6)$ |
| Checkpoint: | $0 \rightarrow 14$ | $14 \rightarrow 9$ | $9$ | |
| Unpacked: | $\{\}$ | $\{5\}$ | $\{6\}$ | |
| References: | ln. 10 | ln. 15,16,18 | ln. 7,8,9 | |

On the other hand, algorithm 1 decomposes it as miloko mithanu na _$(1)$, as it sets the checkpoint behind 'thanu' which avoids that 'thanu na mosa' can be unpacked as a single factor word.

Cause 1: Masked $N(fa * mu)$ has also caused summand words not to get unpacked by algorithm 1, such as in 'veinticinco'. In the case of 'veinticinco', this issue is already resolved with the lines 15-20 of algorithm 2:

'veinticinco' decomposed by advanced algorithm 2:

| $[start{:}end]$: | [0:11] | [6:11] | |
|---|---|---|---|
| Subnumeral: | $N(25)$ | $N(5)$ | |
| Criterion: | $\not< 25/2$ | $\leq \sqrt{25}$ | ⇒veinti_(5) |
| Checkpoint: | $0 \rightarrow 11$ | $11 \rightarrow 6$ | |
| Unpacked: | $\{\}$ | $\{5\}$ | |
| References: | ln. 10 | ln. 15,16 | |

Correct decompositions of $N(21) =$'veintiuno',...,$N(24) =$'veintiquatro' are obtained analogously. However, for $N(20+x)$ with $x > 5$, the summand $x$ is larger than $\sqrt{20+x}$, so it does not satisfy criterion 2. So, when processing 'veintiseis' with algorithm 2, the lines 15-20 do not cause $N(6) =$'seis' to be unpacked. Therefore, we added the else-if clause in line 21, to deal with numerals of order $FA \cdot MU \cdot SU$ with masked $N(fa * mu)$.
Usually, $N(fa * mu)$ is the $substring$ that enters the else-clause in line 10, since it is $> n/2$. However, when $N(fa * mu)$ is masked, it never becomes $substring$. Instead, another $substring$, at least the total $N = N(fa * mu + su)$ itself, will eventually be $> n/2$. Exactly, it will be $substring = FA \cdot MU \cdot SU'$, where $SU'$ is the minimal suffix of $SU$ that makes $FA \cdot MU \cdot SU' = N(fa * mu + su')$ a proper numeral word. Then, after $FA \cdot MU \cdot SU'$ has been processed (ln. 6,7,10), a $substring2$ is browsed for. While the if-clause in line 15 checks unpack criterion 2, the else-if-clause in line 21 checks the remaining arithmetic requirements of criterion 4 for $S = substring2$. Note that the requirement $n(substring2) \mid fa * mu$ is equivalent to $n(substring2) \mid n(substring)$, since $substring = N(fa * mu + substring2)$. In order to use unpack criterion 4, it is still required that $substring2$ has no subnumerals. Therefore, $substring2$ is not immediately unpacked after passing line 21. Instead, it is saved as $maybeUnpack$ (ln. 22). If and only if another $subnumeral2$ is found inside $maybeUnpack$ at a later $start2$ (ln. 14), $maybeUnpack$ is reset again (ln. 19, ln. 22 or ln. 25). If not, then we assume that $maybeUnpack = SU'$ has no subnumerals and unpack it in line 30, as it fulfills the requirements of criterion 4. In this way, the Spanish numeral $N(26) =$'veintiseis' gets decomposed properly by algorithm 2:

```
Language: Spanish     Number: 26 = 20+6
Index:        0           6     9
Numeral:      v e i n t i s e i s
Subnumerals:             |--N(6)-|
Desired decomposition: veinti_(6)
```

'veintiseis' decomposed by advanced algorithm 2:

| $[start{:}end]$: | [0:10] | [6:10] | |
|---|---|---|---|
| Subnumeral: | $N(26)$ | $N(6)$ | |
| Criterion: | $\not< 26/2$ | $< 26/2 \wedge \nmid 26 \wedge$ is atom | ⇒veinti_$(6)$. |
| Checkpoint: | $0 \rightarrow 10$ | $10 \rightarrow 6$ | |
| Unpacked: | $\{\}$ | $\{6\}$ | |
| References: | ln. 10 | ln. 21,22,29,30 | |

It works the same for $N(27)$-$N(29)$.

Also, for most French numerals of the shape 'quatre-vingt-'$\cdot X$, the summand $X$ now gets unpacked properly.

```
Language: French     Number: 91 = 4*20+11
Index:       0           6           12      16
Numeral:     q u a t r e - v i n g t - o n z e
Subnumerals:|----N(4)---| |--N(20)--| |-N(16)-|
Desired decomposition: _-vingt-_(4,11)
```

'quatre-vingt-onze' decomposed by lines 1-37 of advanced algorithm 2:

| $[start{:}end]$: | [0:6] | [7:12] | [0:17] | [13:17] |
|---|---|---|---|---|
| Subnumeral: | $N(4)$ | $N(20)$ | $N(91)$ | $N(11)$ |
| Criterion: | $< 91/2$ | $< 91/2$ | $\not< 91/2$ | $< 91/2 \wedge \nmid 91 \wedge$ is atom |
| Checkpoint: | $0$ | $0$ | $0 \to 17$ | $17 \to 12$ |
| Unpacked: | $\{4\}$ | $\{4, 20\}$ | $\{4, 20\}$ | $\{4, 20, 11\}$ |
| References: | ln. 7,8 | ln. 7,8 | ln. 10 | ln. 21,22,29,30 |

$\Rightarrow$_-_-_$(4, 20, 11)$

However, since criterion 4 is leaky, $X$ can still remain un-unpacked in rare cases where $n(X) \mid n($'quatre-vingt-'$\cdot X)$:

```
Language: French     Number: 96 = 4*20+16
Index:       0           6           12        17
Numeral:     q u a t r e - v i n g t - s e i z e
Subnumerals:|----N(4)---| |--N(20)--| |--N(16)--|
Desired decomposition: _-vingt-_(4,16)
```

'quatre-vingt-seize' decomposed by advanced algorithm 2:

| $[start{:}end]$: | [0:6] | [7:12] | [0:18] | [13:18] |
|---|---|---|---|---|
| Subnumeral: | $N(4)$ | $N(20)$ | $N(96)$ | $N(16)$ |
| Criterion: | $< 96/2$ | $< 96/2$ | $\not< 96/2$ | $\not\leq \sqrt{96}$ and $\mid 96$ |
| Checkpoint: | $0$ | $0$ | $0 \to 18$ | $18$ |
| Unpacked: | $\{4\}$ | $\{4, 20\}$ | $\{4, 20\}$ | $\{4, 20\}$ |
| References: | ln. 7,8 | ln. 7,8 | ln. 10 | ln. 24 |

$\Rightarrow$_-_-seize$(4, 20)$

The same issue also appears for 'quatre-vingt-dix', which is decomposed _-_-dix$(4, 20)$. A related issue appears with the numerals $N(97)$, $N(98)$ and $N(99)$, as their summand begins with the sub-subnumeral $SU' =$'dix'$= N(10)$ and $N(SU') \mid N($'quatre-vingt-'$\cdot SU')$.

```
Language: French     Number: 99 = 4*20+19
Index:       0           6           12      16      20
Numeral:     q u a t r e - v i n g t - d i x - n e u f
Subnumerals:|----N(4)---| |--N(20)--| |-----N(19)-----|
            |-------------N(90)-----------| |--N(9)-|
                                      |N(10)|
Desired decomposition: _-vingt-_(4,19)
```

'quatre-vingt-dix-neuf' decomposed by advanced algorithm 2:

| $[start{:}end]$: | [0:6] | [7:12] | [0:16] | [13:16] | [17:21] |
|---|---|---|---|---|---|
| Subnumeral: | $N(4)$ | $N(20)$ | $N(90)$ | $N(10)$ | $N(9)$ |
| Criterion: | $< 99/2$ | $< 99/2$ | $\not< 99/2$ | $\not\leq \sqrt{99}$ and $\mid 90$ | $< 99/2$ |
| Checkpoint: | $0$ | $0$ | $0 \to 16$ | $16$ | $16$ |
| Unpacked: | $\{4\}$ | $\{4, 20\}$ | $\{4, 20\}$ | $\{4, 20\}$ | $\{4, 20, 9\}$ |
| References: | ln. 7,8 | ln. 7,8 | ln. 10 | ln. 24 | ln. 7,8 |

$\Rightarrow$_-_-dix-_$(4, 20, 9)$

This far we have fixed proper unpacking of summand words. In the few rare cases in French where summands are not unpacked, some lexicon efficiency is lost due to the leakiness of unpack criterion 4. Specifically, the five numerals $N(90)$ and $N(96)$-$N(99)$ cannot be covered by the function _-vingt-_, but need their own seperate lexicon entries.

### 5.3.2 Dealing with overgeneralization

As can be seen in the Examples of 'mak´umi matatu na zinai', 'kaksisataayksi' and 'quatre-vingt-dix': Whenever $N(mu)$ ends before $N(fa * mu)$ (like 'k´umi' in 'mak´umi matatu', 'sata' in 'kaksisataa' and 'vingt' in 'quatre-vingts'), then algorithm 1 unpacks the multiplicator despite our intention. This leads to decompositions like $80 =$_-_s$(4, 20)$ or $201 =$__a_$(2, 100)$. Such functions are prone to overgeneralization. Specifically, '__a_' cannot only generate Suomi numerals between $10^2$ and $10^3$, but also numerals between $10^6$ and $10^{12}$ when the second input is $N(10^6) =$'miljoona' or $N(10^9) =$'miljardi' instead of $N(100) =$'sata'. This is an undue overgeneralization with respect to Objective 2, since '__a_' cannot work with an affine linear function in this value range. Algorithm 2 overcomes this issue with the added lines 38-60. There, it looks for clues to detect an unpacked $MU$ in order to un-unpack it.
If we have 3 unpacked subnumerals $U_1, U_2, U_3$ in a numeral $X$ and they happen to satisfy $u_1 * u_2 + u_3 = x$ with $u_1 < u_2$, then we suspect $MU = U_2, FA = U_1$ and $SU = U_3$, so we would un-unpack $U_2$. Note that first one would need to find the distribution of the unpacked subnumerals on the roles $FA, MU$ and $SU$. Here, it helps to note that—based on Axioms 2 and 3—$MU$ would always have a larger value than the others.
So, if we have 3 unpacked subnumerals, our strategy is: Set the value-largest unpacked subnumeral $U_{max}$ to $maybeMU$ (ln. 38), as we suspect, it may be $MU$. If the other two $U_1, U_2$ satisfy $u_1 * u_{max} + u_2 = x$ (ln. 47) or $u_2 * u_{max} + u_1 = x$ (ln. 49), then we un-unpack $maybeMU = U_{max}$ (ln. 48,50), as the suspicion $U_{max} = MU$ has been heightened.
If we have a number of unpacked subnumerals different from $3$ we use a similar strategy, as can be seen in the lines 40-45 and 52-59 of Algorithm 2.
With this fix, we were able to solve the errors of the sort $MU$ unpacked on a large scale.
For example, it solved the issue in the decomposition of 'kaksisataayksi'$= 201 =$ _ _a_$(2, 100, 1)$, as the algorithm can find out that $2 * 100 + 1 = 201$ and detect $100$ as a multiplicator. By un-unpacking $N(100)$ the desired decomposition $201 =$ _ sataa_$(2, 1)$ is obtained. Likewise, it works in the cases of 'quatre-vingts' and 'mak´umi matatu na zinai'.

| Numeral | $N(80) =$quatre-vingt | $N(34) =$mak´umi matatu na zinai |
|---|---|---|
| | ⇓ | ⇓ |
| Alg. 2 dec. till l. 37 | _-_s$(4, 20)$ | ma_ ma_ na zi_$(10, 3, 4)$ |
| | ⇓ | ⇓ |
| Finding | $80 = 4 * 20$ | $34 = 10 * 3 + 4$ |
| | ⇓ | ⇓ |
| Diagnosis | $N(20)$ is $MU$ | $N(10)$ is $MU$ |
| | ⇓ | ⇓ |
| Alg. 2 final dec. | _-vingts$(4)$ | mak´umi ma_ na zi_$(3, 4)$ |

If in the processing of 'quatre-vingt-seize' algorithm 2 would have unpacked the summand $N(16) =$'seize' properly, then it would also un-unpack the multiplicator $N(20) =$'vingt' after noticing that $96 = 4 * 20 + 16$. This would have led to the desired decomposition $96 =$_-vingt-_$(4, 16)$. Since 'seize' did not get unpacked, the algorithm only checks whether or not $96 = 4 * 20$ or $96 = 4 + 20$ and does not notice that $96 = 4 * 20 + 16$. By lacking this clue, 'vingt' remains unpacked despite our intention. The same happened for 'quatre-vingt-dix' and the 'quatre-vingt-dix$\cdot N(y)$' for $y \in \{7, 8, 9\}$'. However, the French numerals $N(81) - N(89)$ and $N(91) - N(96)$ are properly decomposed by algorithm 2 as _-vingt-_(4,_).
While many errors got fixed by lines 38-60, a few were caused.
In Sakha[6], $N(3) =$'үс' is accidentally contained in $N(100) =$'сүүс' , so in $N(299)$ it gets unpacked among $N(2)$ and $N(99)$ due to its small value. Since $3 * 99 + 2 = 299$, there is the suspicion that $N(99)$ is a multiplicator, so it is un-unpacked and the final decomposition is $299 =$ _ сү_ тоҕус уон тоҕус$(2, 3)$.

[6] In the code, the language is called Yakut

```
Language: Sakha        Number: 299 = 2*100+99
Index:          0     4       9    12   15     19   22  24
Numeral:        и к к и  с ү ү с  т о ҕ у с  у о н  т о ҕ у с
Subnumerals:   |--N(2)-| |-N(100)| |---N(9)--| |N(10)| |---N(9))-|
               |------N(200)-----| |-------N(90)-----|       |N(3|
                         |N(3|        |N(3| |-------N(19)-----|
               |-----------N(209)------------|
                     |-------N(109)------|
               |---------------N(290)-----------------|
                     |----------N(190)-----------|
                           |------N(90)------|
                              |---N(30)---|
                     |-----------------N(199)-----------------|
                           |------------N(99)------------|
                              |---------N(39)---------|
                                   |------N(19)------|
Desired decomposition: _ сүүс _(2,99)
```

'икки сүүс тоҕус уон тоҕус' `decomposed by advanced algorithm 2:`

| $[start{:}end]$: | [0:4] | [0:9] | [5:9] | [7:9] | [10:15] | ... | [10:25] |
|---|---|---|---|---|---|---|---|
| Subnumeral: | $N(2)$ | $N(200)$ | $N(100)$ | $N(3)$ | $N(9)$ | | $N(99)$ |
| Criterion: | $< \frac{299}{2}$ | $\not< \frac{299}{2}$ | $\not\leq \sqrt{299}$ | $\leq \sqrt{299}$ | $< \frac{299}{2}$ | | $< \frac{299}{2}$ |
| Checkpoint: | 0 | $0 \to 9$ | 9 | $9 \to 7$ | 7 | 7 | 7 |
| Unpacked: | $\{2\}$ | $\{2\}$ | $\{2\}$ | $\{2,3\}$ | $\{2,3,9\}$ | ... | $\{2,3,99\}$ |
| References: | 7,8 | 10,11 | 24 | 15,16,18 | 7,8 | 7,8,9 | 7,8,9 |

$\Rightarrow$_ сү_ _$(2,3,99)$
$\Rightarrow$ `Diagnosis: 299 = 3*99+2`
$\Rightarrow$ `N(99) is MU`
$\Rightarrow$_ сү_ тоҕус уон тоҕус$(2,3)$

The accidental appearance of the small-valued numeral 'үс' in 'сүүс' (100) has caused this error. Similar accidents happened in 4 other languages: Breton, Rapa-Nui, Tok-Pisin and Lachixio-Zapotec.
Note that these errors do only minor damage, as they only require one extra lexicon entry for the single incorrectly decomposed numeral.

# 6 Evaluation

We evaluated our algorithm by testing it on data sets of numerals and analyzing the produced output lexica. The data sets are described in Subsection 6.1. The data set of English numerals $< 1000$ induced the following lexicon:

| | | | |
|---|---|---|---|
| one: | $() \to 1$ | two: | $() \to 2$ |
| three: | $() \to 3$ | four: | $() \to 4$ |
| five: | $() \to 5$ | six: | $() \to 6$ |
| seven: | $() \to 7$ | eight: | $() \to 8$ |
| nine: | $() \to 9$ | ten: | $() \to 10$ |
| eleven: | $() \to 11$ | twelve: | $() \to 12$ |
| thirteen: | $() \to 13$ | _teen: | $(x) \to x + 10$ |
| fifteen: | $() \to 15$ | _een: | $(x) \to 18$ |
| twenty: | $() \to 20$ | twenty-_: | $(x) \to x + 20$ |
| thirty: | $() \to 30$ | thirty-_: | $(x) \to x + 30$ |
| forty: | $() \to 40$ | forty-_: | $(x) \to x + 40$ |
| fifty: | $() \to 50$ | fifty-_: | $(x) \to x + 50$ |
| _ty: | $(x) \to 10 * x$ | _ty-_: | $(x, y) \to 10 * x + y$ |
| _y: | $(x) \to 80$ | _y-_: | $(x, y) \to 10 * x + y$ |
| _ hundred: | $(x) \to 100 * x$ | _ hundred and _: | $(x, y) \to 100 * x + y$ |

.

We have left out the domains of the functions to spare space. All functional equations are correct (Objective 2) and the lexicon comprises 30 entries (Objective 1). The performance appears human-like. A broader and more comparative evaluation follows in the upcoming subsections. In subsection 6.2 we compare three induced grammars with expert-made gold standards. In subsection 6.3 we analyze the correctness of all induced grammars (Objective 2) and in subsection 6.4 the compactness (Objective 1).

## 6.1 Data

From languagesandnumbers.com we obtained dictionaries of number-numeral pairs for numbers up to 1000 in over 250 languages, unless a language did not deliver numerals till 1000.
From the Python package num2words, we got a dictionary of number-numeral pairs for numbers up to 1000 and a sample of 4-digit and 5-digit numbers in over 30 languages. The sample contains the number 27206 and all 4- and 5-digit numbers that we could reasonably imagine to be contained in $N(27206)$ as a subnumeral in a base 10 system, which are

$$1002, 1006, 1100, 1200, 1206, 7000, 7002, 7006, 7100, 7200, 7206, 10000, 17000, \\ 17200, 17206, 20000, 27000, 27006 \text{ and } 27200. \tag{3}$$

In base 20 or other base X systems, other subnumerals would be conceivable, but all base 20 system languages that we have in our data base either transition into base 10 when numbers become bigger, or the data base does not have numerals for numbers over 1000.
Using the TeX code from Derzhanski and Veneva (2020), we generated Birom numerals till 120 and Yoruba numerals till 184.
All data sets only contain the standard grammatical forms of the numerals, since we assume that any challenge that an alternate form may pose on the performance of the numeral decomposer comes up in an analogous form in another language. The interested reader may challenge this claim by testing the decomposer published on GitHub, see I. K. Maier, 2023.

## 6.2 Comparison with Expert-Made Grammars

In this subsection, we compare three expert-made grammars with their numeral-decomposer-induced counterparts. Derzhanski and Veneva (2020) present TeX implementations of grammars for Bulgarian numerals till 99, Birom numerals till 120 and Yoruba numerals till 184. The grammars come from solutions of exercises of the International Linguistic Olympiad and other linguistic contests. The authors chose Bulgarian, Birom and Yoruba because their numeral systems offer a diverse variety of features. In particular, Bulgarian uses a standard base 10 system with subnumeral order factor-multiplicator-summand, Birom uses a base 12 system involving backward counting and order multiplicator-factor-summand, and Yoruba uses a combination of base 20 and base 10 with even more backward counting and order summand-multiplicator-factor.
Table 1 shows a direct comparison of induced and expert-made grammar for Birom. Comparisons for Bulgarian and Yoruba can be found in Appendix A. Notably, for Bulgarian both numeral decomposer version induced the same grammar as the experts, so they worked perfectly.
In order to evaluate the other comparisons, we calculate the *accuracy* value that Hammarström (2008) used. He reviews grammars as clusterings, where each cluster is the set of expressions generated by one rule. If $I$ is the set of induced clusters and $G$ the set of gold standard clusters, then accuracy is defined as

$$Acc = \frac{\frac{1}{|I|}\sum_{r\in I} prec(r,G) + \frac{1}{|G|}\sum_{r\in G} prec(r,I)}{2}, \text{ with}$$
$$prec(r,X) = \frac{|r| - |\{r_x \in X \mid r \cap r_x \neq \emptyset\}| + 1}{|r|}.$$

The formula is similar to cluster purity, see Manning et al., 2008, chapter 16.3. While purity measures how much of one induced rule can be covered by one gold rule, accuracy measures how many gold rules it takes to cover one induced rule completely.

<table>
<tr><th colspan="2">Induced grammar</th><th colspan="2">Expert-made grammar</th></tr>
<tr><th>Rule/Function</th><th>Values</th><th>Rule/Function</th><th>Values</th></tr>
<tr><td>ATOMS</td><td>1–8,12</td><td>ATOMS</td><td>1–8,12</td></tr>
<tr><td>Sāā_<br>(x)$\mapsto$ -1x+12</td><td>9–11</td><td>Sāā_<br>(x)$\mapsto$ -1x+12</td><td>9–11</td></tr>
<tr><td>bākūrū bī_<br>(x)$\mapsto$ 12x</td><td>$12x$ for<br>$x \in \{2–8\}$</td><td>bākūrū bī_<br>(x)$\mapsto$ 12x</td><td>$12x$ for<br>$x \in \{2–8\}$</td></tr>
<tr><td>bā_ Sāābī_<br>(x,y)$\mapsto$ 12x-12y</td><td>108, 120</td><td>bākūrū Sāābī_<br>(x)$\mapsto$ (12-x)·12</td><td>108, 120</td></tr>
<tr><td>kūrū na gwĒ _<br>(x)$\mapsto$ 13</td><td>13</td><td rowspan="3">_ na gwĒ _<br>(x,y)$\mapsto$ x+y</td><td rowspan="3">$12x$+1 for<br>$x \in \{1–9\}$</td></tr>
<tr><td>bākūrū bī_ na gwĒ _<br>(x,y)$\mapsto$ 12x+y</td><td>$12x$+1 for<br>$x \in \{2–8\}$</td></tr>
<tr><td>bā_ Sāābī_ na gwĒ _<br>(x,y,z)$\mapsto$ 109</td><td>109</td></tr>
<tr><td>kūrū na vE_<br>(x)$\mapsto$ x+12</td><td>14–23</td><td rowspan="3">_ na vE_<br>(x,y)$\mapsto$ x+y</td><td rowspan="3">$12x$+$y$ for<br>$x \in \{1–9\}$<br>$y \in \{2–11\}$</td></tr>
<tr><td>bākūrū bī_ na vE_<br>(x,y)$\mapsto$ 12x+y</td><td>$12x$+$y$ for<br>$x \in \{2–8\}$<br>$y \in \{2–11\}$</td></tr>
<tr><td>bā_ Sāābī_ na vE_<br>(x,y,z)$\mapsto$ 8x+4y+z</td><td>110–119</td></tr>
</table>

Table 1: An advanced-numeral-decomposer induced grammar for Birom language in comparison with an expert-made gold standard. The decomposer has about the same idea as the expert but it splits up '_na gwĒ _' and '_ na vE_' in three functions to avoid overgeneralization.

| **Accuracies** | Bulgarian | Birom | Yoruba |
|---|---|---|---|
| Advanced | 100% | 99.13% | 98.73% |
| Prototype | 100% | 89.28% | 89.06% |

Table 2: Accuracy values (Hammarström, 2008) of grammars induced by prototype and advanced numeral decomposer in relation to Derzhanski and Veneva (2020)'s expert-made grammars. For Bulgarian, both decomposer versions induced the same grammar as the experts made.

For the Birom induced grammar, we calculate an accuracy of $99.13$%. All induced rules can be covered with one single expert-made rule, so $prec(r, G) = 1$ for all $r \in I$. Out of the $r \in G$, the 9 atoms and the 3 functions 'Sāā_', 'bākūrū bī_' and 'bākūrū Sāābī_' are covered by one single induced rule each, while '_ na gwĒ _' and '_ na vE_' are covered by 3 rules. Thus

$$Acc_{Birom} = \frac{1 + \frac{1}{14}(12 \cdot 1 + \frac{9-3+1}{9} + \frac{90-3+1}{90})}{2} = 0.9913.$$

The accuracies of the other induced grammars can be found in Table 2. For comparison, the numeral grammars induced by Hammarström (2008) had an average accuracy of $71.56$% and a median accuracy of $90$%.

## 6.3 Regarding Correct Functional Equations

For the induced grammars, the functional equation of each template is calculated by affine linear regression. A functional equation of a template is correct when it computes the correct number value for each numeral generated by the template. In this subsection we report about every incorrect functional equation occurring in our data.

We note that apart from our data, which do not go beyond $10^6$, big English numeral words like 'trillion', 'quadrillion', 'quintillion' etc. are related to the Latin numerals 'tria', 'quattuor', 'quinque', while the impact of these implied subnumerals is not linear but exponential.

Inside our data, we have summarized a report regarding Objective 2 in the following table. Not all errors are caused by bad decomposition. Some occurred due to unintuitive context sensitivity, which we will explain later.

| Error causes | Languages |
|---|---|
| Bad decomposition | 3: Tongan, Kiribati, Nyungwe |
| Context sensitivity | 4: Choapan-Zapotec, Nume, Farsi, Hebrew |
| Incorrect input data(?) | 7: Haida, Purepecha, Susu, Dogrib, Tunica, Yao, Yupik |
| No errors | 239: the rest |

In 239 out of 253 languages, the advanced numeral decomposer did not do any undue generalizations. So, for each template functions in all 239 languages, an exact affine linear equation was found that interprets all its output numerals with the correct number value.
In the 14 remaining languages, undue generalizations led to inexact functional equations and thus incorrect interpretations of numerals.
In 11 out of the 14 failed languages, we consider the wrong interpretation reasonable enough that humans could misinterpret them as well.
In many of these cases we suspect that the data from languagesandnumbers.com have errors: In 6 languages, Purepecha, Susu, Dogrib, Tunica, Yao and Yupik, we found pairs of numbers with exact same numeral. It is also concievable that these pairs actually differ in intonation or something, and the differences are just not visible in the delivered written form. We also suspect wrong data in Haida, which we explain later.
In the other 5 languages out of the 11, context sensitivities led to errors, i.e., there are compound numerals $X \cdot Y$ and $X' \cdot Y'$, where $(X, X')$ and $(Y, Y')$ are pairs of intuitively similar numerals but the calculation of $n(X \cdot Y)$ out of $x$ and $y$ is fundamentally different from the calculation of $n(X' \cdot Y')$ out of $x'$ and $y'$:
Choapan-Zapotec: While $N(1) =$ 'tu', $N(2) =$ 'chopa' and $N(3) =$ 'tzona', and 'chopa galo' and 'tzona galo' mean $2 * 20$ and $3 * 20$ respectively, the numeral 'tu galo' means $20 - 1$ instead of $1 * 20$.
Nume: When a 1-digit numeral $S$ (in base 10) is affixed to 'muweldul ', then it means $100 * s$, but if $S$ is a 2-digit numeral, then is means $100 + s$.
Farsi (Persian): In the Latin-transcripted form, we have $N(600) =$ 'sheshsad', composed as $N(6)\cdot$'sad'. The numeral $N(300)$ is similarly composed, but $N(3) =$ 'se' gets inflected to 'si', which accidentally is $N(30)$, so we have a template _sad mapping $6$ to $600$ and $30$ to $300$. Actually, an affine linear equation $x \mapsto 600 + \frac{300-600}{30-6} * (x - 6)$ is still construable, but for code efficiency reasons we have only allowed integer coefficients for the functional equations.
Haida: While $N(2) =$ 'sdáng', $N(3) =$ 'hlgúnahl' and $N(8) =$ 'sdáansaangaa', and 'lagwa uu sdáng' and 'lagwa uu hlgúnahl' mean $2 * 20$ and $3 * 20$ respectively, according to languagesandnumbers.com the numeral 'lagwa uu sdáansaangaa' means $80$ instead of $8 * 20$. We suspect that this is wrong information, since according to omniglot.com $80$ means 'lagwa uu stánsang', which is logical since 'stánsang' $= N(4)$.
Hebrew: While $N(3) =$ שלש, $N(4) =$ ארבע and $N(10) =$ עשר, and $3 * 10$ and $4 * 10$ are written שלשים and ארבעים, respectively, the numeral עשרים means $10 + 10$ instead of $10 * 10$.
In the 3 remaining languages, undue generalizations were made due to bad decompositions:
In Tongan-Telephone-Style, numbers are—with some minor inflections—simply called by the sequence of their decimal digits. The total lack of multiplicator words leads to various words being identified with the template _ _. This template can sometimes mean $(x_0, x_1) \mapsto 10 * x_0 + x_1$ for 2-digit numerals and sometimes $(x_0, x_1) \mapsto 100 * x_0 + x_1$. The fact that the rough value size of a numeral cannot be instantaneously estimated during reading—as any further amount of digits could still be added— also makes it impossible to make proper use of the seperated $start2$ loop in the algorithm. One could argue that this language does not really follow Hurford's Theory of Numerals. This take can be justified by an argument that the development of this numeral language is more influenced by telecommunication technology than by nature, so those numerals may not be considered a natural part of a language.
In Gilbertese (Kiribati), the numerals $N(90)$ and $N(900)$ can accidentally(?) be presented as ru$\cdot N(40)$ and ru$\cdot N(400)$. This causes the numerals $N(90 + s)$ and $N(900 + s)$ to be decomposed ru_$(40 + s)$ and ru_$(400 + s)$ respectively. A unification of these ru_ has no proper affine linear equation, since the points $(41, 91), (42, 92), (401, 901)$ do not lie on a straight line.
In Nyungwe, again multiplicators got unpacked and generalized. The numerals $N(31), N(41), N(301)$ and $N(401)$ got all identified with the template ma_ ma_ na ibodzi with the inputs $(10, 3), (10, 4), (100, 3)$ and $(100, 4)$ respectively. As these input-output combinations do not lie on a straight surface, an exact affine linear functional equation does not exist for the template ma_ ma_ na ibodzi.

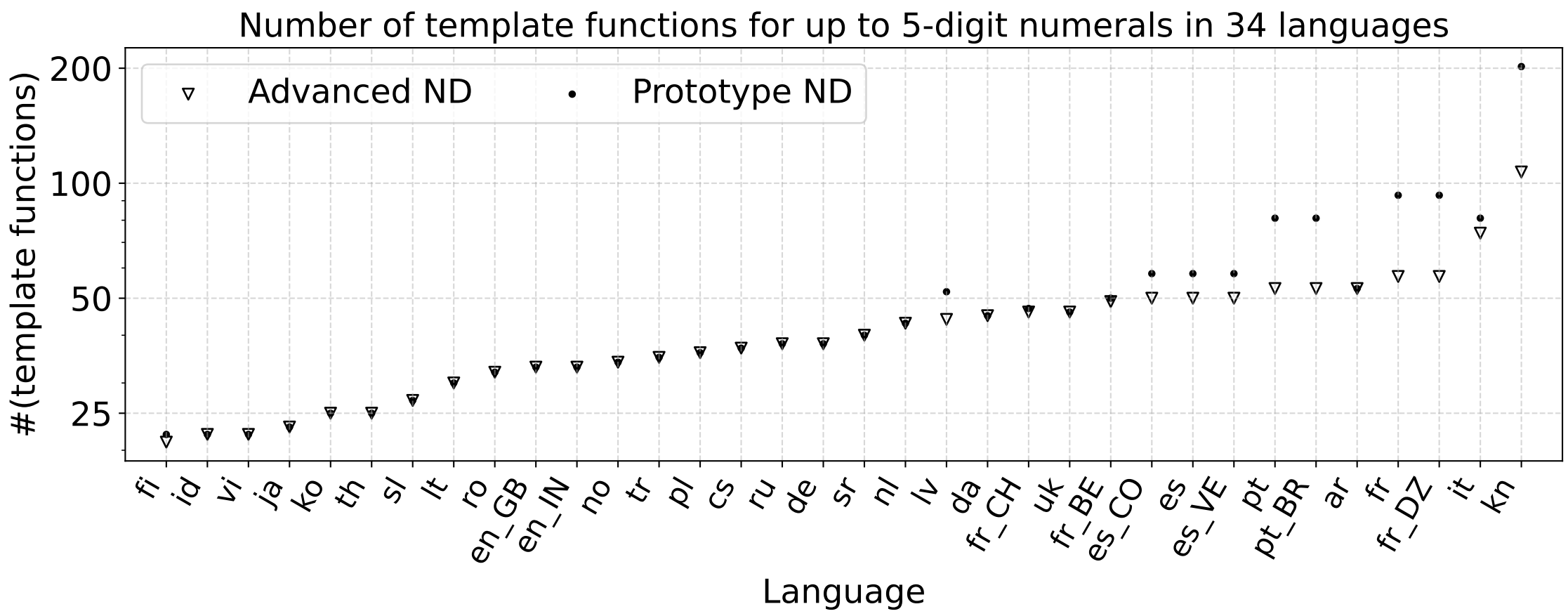

Figure 3: Sizes of grammars induced by numeral decomposer versions for numerals of numbers 1-999 and the numbers in Equation 3.

## 6.4 Regarding Lexicon Sizes

In this subsection, we discuss the lexicon sizes of numeral-decomposer induced grammars, which according to Objective 1 should be as small as possible.
Figure 3 shows in 34 languages how many different template functions the two numeral decomposer versions induce to cover the numerals from 1 to 999 plus the sample of 4 to 5-digit numbers (Eq. 3). The languages are sorted by the $y$ values of the advanced version, so one can conclude that, e.g., in 24 of the 34 languages, the advanced numeral decomposer covers the numerals in 50 templates or less. Figure 4 shows the number of induced templates for numerals up to 999 and for up to 399. For these ranges of values, we have data from over 200 languages. Therefore, the $x$ axis does not show the names of the languages, but it represents their ordinal positions with respect to the number of advanced-numeral-decomposer-induced template functions. The plots imply that the advanced numeral decomposer induces compact numeral grammars in most languages. In 195 languages out of 234, it maps the numerals till 1000 to 50 templates or less. In only 8 languages, it produces over 100 different templates, which are Bavarian, Makhuwa, Hayastani (Armenian), Kartvelian (Georgian), Zulu, Timbisha, Xhosa and Kannada (kn). These languages have in common that most subnumerals of compound numerals are masked due to dropping or inflecting their last or first letter(s).
The scattered grey data points above the black curves show that the templates induced by the prototype numeral decomposer are situationally less generalizing than the templates induced by the advanced numeral decomposer.
In order to give a comparison, we conducted grammar induction for numerals till 99 not only with the numeral decomposer versions but also with GITTA (Winters & Raedt, 2020). GITTA is a general tool that induces context-free grammars for natural language input. We assist GITTA by adding spaces into the numerals at any position where a subnumeral begins or ends.
To make the comparison even, we have to convert the numeral template grammars into CFGs. Therefore, each template function $f$ becomes the right-hand side of a context-free rule $S \to f$, where each input slot '_' of $f$ is replaced by a unique nonterminal $N$, which yields a production rule $N \to g$ for each function $g$ that can be applied on the input slot. Nonterminals that produce the same set of right-hand sides are merged. Example: The English function _ty-_:$\{6, 7, 9\} \times \{1, ..., 9\}$ is modeled using 13 context-free rules $S \to A$ty-$B$, $A \to$ six,seven,nine and $B \to$ one,...,nine. The function twenty-_:$\{1, ..., 9\}$ is modeled as $S \to$ twenty-$C$ with $C \to$ one,...,nine. Since $B$ and $C$ produce the same words, they merge, so all rules of $C$ are removed and twenty-$C$ is renamed to twenty-$B$.
Using the present conversion, the numeral-decomposer-induced CFGs do not overgenerated.

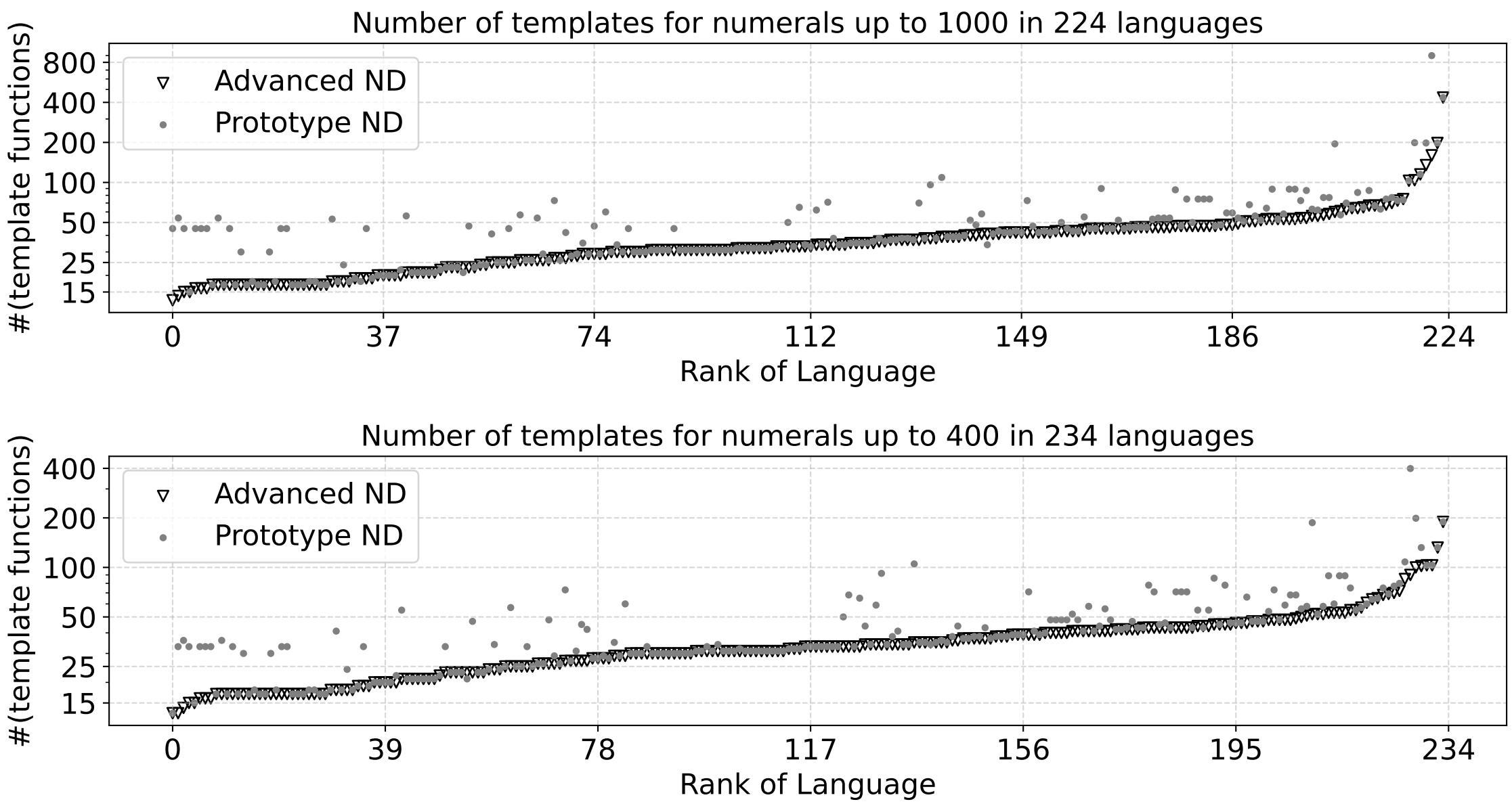

Figure 4: The black data points show in how many languages the advanced numeral decomposer induced less than 100, 50 or 25 different template functions to cover numerals till 1000 (top) or till 400 (bottom). Only about a sixth of the languages got more than 50 templates to get their numerals till 1000 covered. The grey data points show that the prototype decomposer induces much larger grammars in some languages.

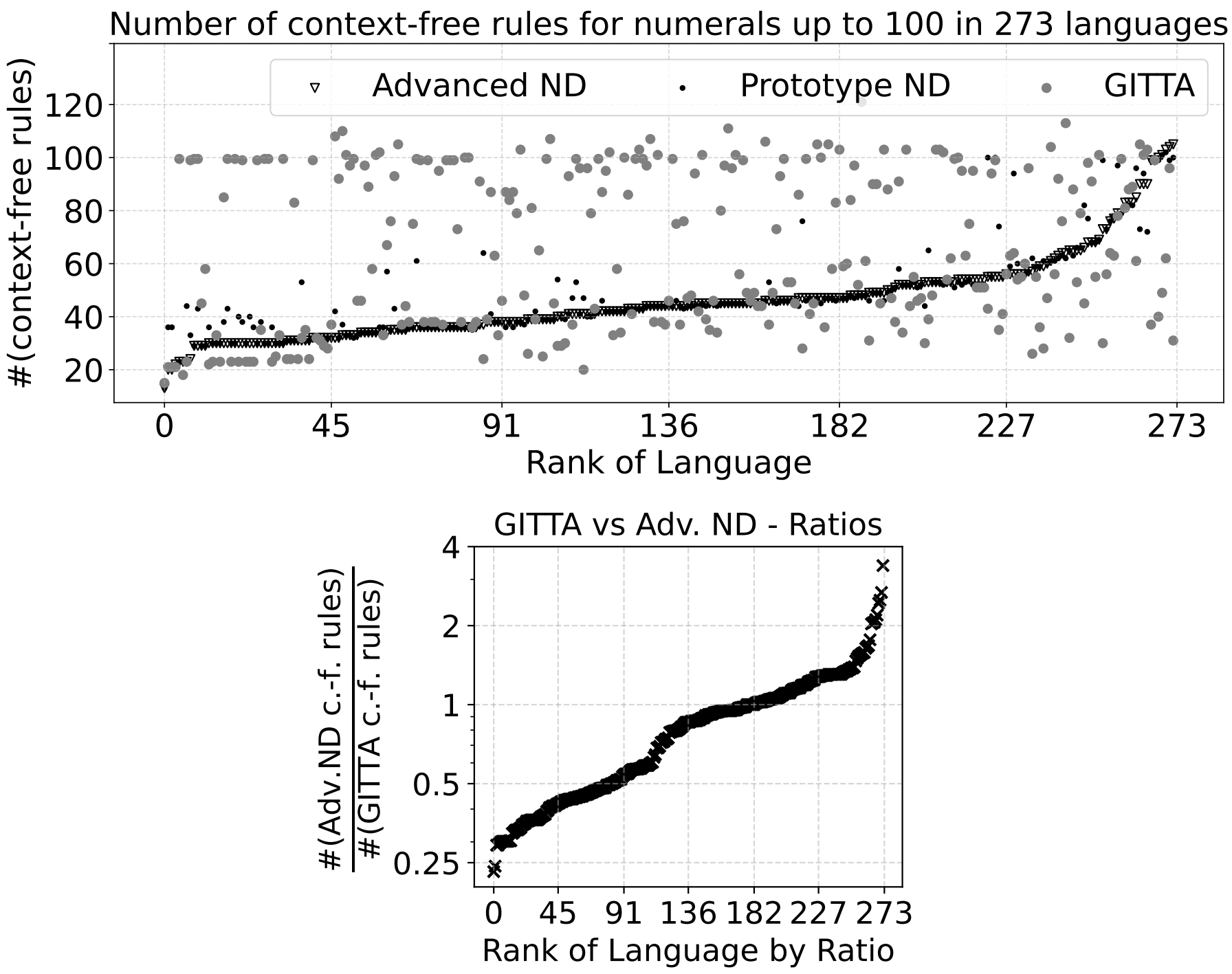

Figure 5: **Context-free-grammar induction comparison between numeral decomposer and GITTA:** The upper plot shows that in about 2 thirds of the languages the advanced numeral decomposer induces exact CFGs for numerals till 100 with less than 50 rules. GITTA's CFGs are mostly larger. The lower plot shows that in about 2 thirds of languages the CFGs induced by the advanced decomposer are smaller than GITTA's CFGs. In about a third of the languages they are smaller than half of GITTA's size.

Therefore, we also made GITTA not overgeneralize by setting the parameter $relative_similarity_threshold$ to $1$.
Figure 5 (upper) shows the number of context-free rules induced for numerals till 99 by GITTA and the two numeral decomposer versions. Both decomposer versions outperform GITTA both on average and on median, as we show in Figure 5 (lower) and in the following table.

| #(Context-free rules) | GITTA | Adv. ND | Proto. ND |
|---|---|---|---|
| Average | 65.95 | 46.20 | 47.77 |
| Median | 61 | 44 | 44 |

In addition, GITTA does not deliver arithmetical attributes to the rules.
However, in some languages, the numeral decomposer does significantly worse than GITTA. This is because GITTA is not asked for arithmetic attributes, so it uses generalizations that the numeral decomposer considers too risky as they might cause generalizations that cannot be covered in with affine linear equations. E.g., in Bavarian, the decomposers induce 101 rules, while GITTA only induces 49. Bavarian 2-digit numerals usually have masked factors and summands, so the numeral decomposer cannot unpack them. However, GITTA can still generalize $N(fa * mu)$ then, while the unpacking criteria forbid the numeral decomposers to do the same as $fa * mu > (fa * mu + su)/2$. GITTA also profits from generalizing empty strings, which the numeral decomposer does not dare.

# 7 Summary

Codes of both algorithm versions are published in I. K. Maier (2023). The list of languages dealt with can also be found there.
The advanced numeral decomposer can induce numeral grammars in a diverse variety of natural languages. In 2 out of 3 cases, its induced CFGs are often more compact than CFGs induced by the state-of-the-art grammar induction algorithm GITTA (Winters & Raedt, 2020). The main limitation of the numeral-decomposer-induced grammars is that they only allow for generalization of entire subnumerals. In languages like Kartvelian Hayastani or Bavarian, numerals often drop or change letters when used as subnumerals, so they cannot be detected and generalized, which significantly enlarges the numeral-decomposer-induced grammars.
In Bulgarian, Birom and Yoruba, we compared numeral-decomposer-induced grammars to expert-made gold standard grammars (Derzhanski & Veneva, 2020). All three induced grammars are similar or equal to the gold standards. Specifically, they yield higher *accuracies* than the grammars that Hammarström's $k$-cluster algorithm had deduced.
The numeral-decomposer-induced grammars have the inherent advantage of parallelly induced arithmetical attributes over general syntactic grammar induction algorithms. In 239 out of 253 languages, the induced arithmetical attributes were entirely correct. Incorrect arithmetic was mainly induced in such numerals where a misunderstanding is perceivable for humans.
Another advantage of the numeral decomposer is that it can decompose numerals incrementally in a learning process. Syntactic grammar induction requires comparisons of expressions, like 'twenty-one', with other expressions of the same abstraction level, like 'twenty-two'..., to find patterns for generalization. In contrast, the numeral decomposer just needs to know the expressions of the lower abstraction level, e.g., when it knows that 'one' is the numeral of $1$, then it understands that 'one' is a generalizable part of 'twenty-one' based on the unpack criteria.
Incrementality facilitates the expansion of existing grammars. Most existing grammar induction methods are non-incremental (Muralidaran et al., 2021).

# 8 Outlook

The tests have only been conducted on grammatical standard forms of numerals. On the other hand, a major limitation of the numeral decomposer is the fact that it cannot unpack and generalize subnumerals that appear in a masked/inflected form. Both shortcomings could be dealt with by letting the numeral decomposer learn several grammatical forms of each numeral. In this way, a stem of all forms could be determined. In the process, the numeral decomposer could not only detect and evaluate fully contained subnumerals, but also just stems of such subnumerals. Depending on the degree of tolerance, it may therefore detect and unpack the 'thir' in 'thirteen' if it has learned before that 'third' is a grammatical variant of $N(3)$.
Such measures could greatly support generalizations. It could even prevent overgeneralizations in cases where the tolerance helps detect $N(fa * mu)$. E.g., when 'quatre-vingt**s**' is detected in 'quatre-vingt-deux' the problem dealt with in Subsection 5.3.2 - which partly persists - does not come up.
On the other hand, it can cause that substrings are unintentionally detected as subnumerals due to a random similarity, which may cause deeper problems.
The numeral decomposer may be tested on learning numerals in a random unchronological order. The test poses the challenge of working with a limited lexicon. When the decomposer gets to learn 'sixteen' before 'six', it cannot detect and unpack 'six', which leads a shortcoming in generalization. The shortcoming could be quantified by learning numerals in a random order that uses a suitable probability distribution. Unchronological learning could also be dealt with by giving up incrementality and decomposing 'sixteen' again after 'six' got learned.
The numeral-decomposer-based incremental learning algorithm could also involve reinforcement. For a learned template like '_ty-_' the learning algorithm could think up words by inserting alternative subnumerals into the slots. It just needs some sort of supervisor that accepts or rejects generalizations of learned words. Such a reinforcement learning offers many possibilities for application-related projects:

- If the supervisor is replaced by a human, the reinforcement learning algorithm can work like a chatbot that can create generative grammars for number words in small languages with human support. The human would only need to answer questions like 'What is the numeral of number $x$?' and 'Does numeral $X$ exist?'.
- If the learner is able to extract numerals out of text data, only answers to questions of the form 'Does numeral $X$ exist?' would be needed. And these questions could be answered with a search engine and a statistical model, which—given a numeral $X$ and the number of search results for $X$—could decide if $X$ is a correctly spelled numeral.

# A Yoruba and Bulgarian induced and expert-made grammars

| Bulgarian induced grammar | | Expert-made grammar | |
|---|---|---|---|
| Rule/Function | Values | Rule/Function | Values |
| ATOMS | 1–12,20 | ATOMS | 1–12,20 |
| _nadeset<br>$(x) \mapsto x + 10$ | 13–19 | _nadeset<br>$(x) \mapsto x + 10$ | 13–19 |
| dvadeset i _<br>$(x) \mapsto x + 20$ | 21–29 | dvadeset i _<br>$(x) \mapsto x + 20$ | 21–29 |
| _deset<br>$(x) \mapsto 10x$ | 10$x$ for<br>$x \in \{3–9\}$ | _deset<br>$(x) \mapsto 10x$ | 10$x$ for<br>$x \in \{3–9\}$ |
| _deset i _<br>$(x, y) \mapsto 10x + y$ | 10$x$+$y$ for<br>$x \in \{3–9\}$<br>$y \in \{1–9\}$ | _deset i _<br>$(x, y) \mapsto 10x + y$ | 10$x$+$y$ for<br>$x \in \{3–9\}$<br>$y \in \{1–9\}$ |

<table>
<tr><th colspan="2">Yoruba induced grammar</th><th colspan="2">Expert-made grammar</th></tr>
<tr><th>Rule/Function</th><th>Productions</th><th>Rule/Function</th><th>Productions</th></tr>
<tr><td>ATOMS</td><td>1–10,20</td><td>ATOMS</td><td>1–10,20</td></tr>
<tr><td>ogun _<br>$(x) \mapsto 20x$</td><td>20$x$ for<br>$x \in \{2–9\}$</td><td>ogun _<br>$(x) \mapsto 20x$</td><td>20$x$ for<br>$x \in \{2–9\}$</td></tr>
<tr><td>_ l-e.wa<br>$(x) \mapsto x + 10$</td><td>11–14</td><td rowspan="3">_ l-_<br>$(x, y) \mapsto x + y$</td><td rowspan="3">10$x$+$y$ for<br>$x \in \{1–18\}$<br>$y \in \{1–4\}$</td></tr>
<tr><td>_ l-ogun<br>$(x) \mapsto x + 20$</td><td>21–24</td></tr>
<tr><td>_ l-ogun eji<br>$(x) \mapsto x + 40$</td><td>41–44</td></tr>
<tr><td>_ dinogun _<br>$(x, y) \mapsto x + 10y$</td><td>30–34</td><td rowspan="2">e.wa din ogun _<br>$(x) \mapsto 20x - 10$</td><td rowspan="2">20$x$-10 for<br>$x \in \{2–9\}$</td></tr>
<tr><td>_ din_ _<br>$(x, y, z) \mapsto x - y + 20z$</td><td>20$x$+$y$ for<br>$x \in \{2–9\}$<br>$y \in \{10–14\}$</td></tr>
<tr><td>_ din ogun<br>$(x) \mapsto -1x + 20$</td><td>15–19</td><td rowspan="4">_ din _<br>$(x, y) \mapsto x - y$</td><td rowspan="4">10$x$-$y$ for<br>$x \in \{2–18\}$<br>$y \in \{1–5\}$</td></tr>
<tr><td>_ din e.wa dinogun _<br>$(x, y) \mapsto 25$</td><td>25</td></tr>
<tr><td>_ din _ dinogun _<br>$(x, y, z) \mapsto -x + 3y$</td><td>26–29</td></tr>
<tr><td>_ din _ din_ _<br>$(x, y, z, a) \mapsto -x - y + 20a$</td><td>20$x$+$y$ for<br>$x \in \{2–9\}$<br>$y \in \{5–9\}$</td></tr>
<tr><td>_ _<br>$(x, y) \mapsto x + 20y - 20$</td><td>20$x$+$y$ for<br>$x \in \{2–9\}$<br>$y \in \{$-5– -1,1–4$\}$</td><td></td><td></td></tr>
</table>